\documentclass[reprint,amsmath,amssymb,aps,superscriptaddress,longbibliography,floatfix,nobalancelastpage]{revtex4-2}

\usepackage[T1]{fontenc}
\usepackage[utf8]{inputenc}
\usepackage{lmodern}
\usepackage[protrusion=true,expansion=true]{microtype}
\usepackage{graphicx}
\usepackage{multirow}
\usepackage{dcolumn}
\usepackage{bm}

\newcommand{\backhead}[1]{\par\addvspace{\medskipamount}\noindent\textbf{#1}\par\nobreak\smallskip\nobreak}
\newcommand{\cstk}[2]{\begin{tabular}[t]{@{}c@{}}#1\\#2\end{tabular}}
\newcommand{\hy}{-\penalty0\hskip0pt\relax}
\newcommand{\ms}[2]{\begin{tabular}[c]{@{}c@{}}#1\\$\pm$\,#2\end{tabular}}
\newcommand{\stk}[2]{\begin{tabular}[t]{@{}l@{}}#1\\#2\end{tabular}}
\usepackage[colorlinks=true,linkcolor=blue,citecolor=blue,urlcolor=blue]{hyperref}
\hypersetup{
  pdftitle={Lightweight Deep Learning Framework for Astrocyte-Centric 3D Reconstruction across Normal and Stress-associated Ultrastructural States},
  pdfauthor={Bowen Deng, Gwang-Jin Jeong, Seulgi Noh, Ji Young Mun, Na Young Kim}}

\begin{document}
\setlength{\emergencystretch}{1.5em}
\renewcommand{\topfraction}{0.95}
\renewcommand{\textfraction}{0.05}
\AtBeginDocument{\addtolength{\baselineskip}{0pt plus 0.4pt}}
\makeatletter\@clubpenalty=10000 \clubpenalty=10000 \widowpenalty=10000\makeatother

\renewcommand{\bibsection}{\section*{References}}
\bibpunct{}{}{,}{s}{}{,}
\renewcommand{\thetable}{\arabic{table}}
\makeatletter
\def\fnum@figure{\textbf{\figurename~\thefigure}}
\def\fnum@table{\textbf{\tablename~\thetable}}
\makeatother

\title{Lightweight Deep Learning Framework for Astrocyte-Centric 3D Reconstruction across Normal and Stress-associated Ultrastructural States}

\author{Bowen Deng}
\email{b28deng@uwaterloo.ca}
\affiliation{Institute for Quantum Computing, Department of Electrical and Computer Engineering, University of Waterloo, Waterloo, Canada}

\author{Gwang-Jin Jeong}
\affiliation{Department of Neuroscience, Korea Brain Research Institute, Daegu, Republic of Korea}
\affiliation{Department of New Biology, Daegu Gyeongbuk Institute of Science and Technology, Daegu, Republic of Korea}

\author{Seulgi Noh}
\affiliation{Department of Neuroscience, Korea Brain Research Institute, Daegu, Republic of Korea}

\author{Ji Young Mun}
\email{jymun@kbri.re.kr}
\affiliation{Department of Neuroscience, Korea Brain Research Institute, Daegu, Republic of Korea}

\author{Na Young Kim}
\email{nayoung.kim@uwaterloo.ca}
\affiliation{Institute for Quantum Computing, Department of Electrical and Computer Engineering, University of Waterloo, Waterloo, Canada}
\affiliation{Waterloo Institute for Nanotechnology, University of Waterloo, Waterloo, Canada}

\date[]{}

\begin{abstract}
The morphology of astrocytes is closely associated with their ability to support and modulate neural circuit functions; however, its reconstruction from volume electron microscopy (vEM) data remains labor\hy{}intensive and computationally resource\hy{}intensive, placing substantial barriers on its broader adoption and scalability despite the high spatial resolution of vEM. We present AstroSeg-vEM, a lightweight, astrocyte\hy{}centric framework for morphology\hy{}preserving astrocyte segmentation, 3D reconstruction, and quantitative morphological analysis in vEM images. AstroSeg-vEM combines transformer\hy{}based 2D segmentation with dedicated post\hy{}processing methods and operates directly on high\hy{}resolution EM sections, making it well suited to anisotropic vEM data and downstream proofreading workflows while preserving the full in\hy{}plane resolution of the original EM data. Across four mouse hippocampal datasets representing normal and morphologically defined stress\hy{}associated astrocyte ultrastructural states, AstroSeg-vEM demonstrates robust performance and generalization, achieving average Dice and weighted Dice values of 0.92 and 0.95, respectively, on a held\hy{}out test dataset. Its lightweight architecture enables efficient, high\hy{}fidelity reconstruction on consumer\hy{}grade hardware, providing a scalable and accessible platform for large\hy{}scale astrocyte ultrastructure analysis and future astrocyte\hy{}neuron interaction studies.
\end{abstract}

\maketitle

\section{Introduction}

Astrocytes are essential regulators of synaptic function, neural circuit homeostasis, and neurovascular coupling in the hippocampus\cite{ref1,ref2,ref3}. Their fine perisynaptic and perivascular processes form highly ramified, spatially interwoven territories that interact with dendrites, synapses, and blood vessels. Notably, the astrocytic processes\cite{ref4,ref5,ref6} from specialized nano\hy{}domains mediate neurotransmitter uptake, ion buffering, lipid trafficking, and secretion of antioxidants and trophic factors that support neuronal survival and plasticity\cite{ref7}; however, these structures are beyond the resolution of light microscopy\cite{ref6} and quantitative analysis remains technically challenging\cite{ref8,ref9}. Astrocyte structural heterogeneity\cite{ref10}, defined here as nanoscale changes in irregular process organization and spatial coverage within the synaptic microenvironment\cite{ref11,ref12}, is especially difficult to quantify in 3D analysis. Hence, high\hy{}resolution volume electron microscopy (vEM) is essential for ultrastructural analysis, enabling direct visualization of how astrocytes organize the local synaptic microenvironment, including perisynaptic process geometry, synaptic apposition, and spatial occupancy. Most astrocyte vEM studies still rely on manual or semi\hy{}automated segmentation and proofreading\cite{ref13,ref14,ref15,ref16,ref17,ref18,ref19}, making reconstruction labor\hy{}intensive, operator\hy{}dependent, difficult to scale, and susceptible to split and merge errors. These challenges are particularly problematic for astrocytes because their thin, fragmented, and highly interconnected processes occupy only a small fraction of each section while extending through dense ultrastructures containing dendrites, axons, synapses, and vascular structures. Thus, scalable and reproducible astrocyte\hy{}centered segmentation remains a major bottleneck for quantitative analysis of astrocyte morphology and cell--cell interactions in vEM datasets.

Recent advances in deep learning have substantially improved EM image segmentation and have provided greater automation of downstream reconstruction workflows. Fully 3D segmentation architectures have demonstrated strong performance in vEM segmentation, especially in large\hy{}scale neuron\hy{}centric connectomics\cite{ref20}. While achieving strong segmentation performance, such 3D approaches are computationally demanding and typically require high\hy{}memory GPUs (\textgreater{} 16 GB VRAM), hampering scalability and throughput. For astrocyte segmentation in vEM, 3D deep learning\hy{}based pipelines have also been applied to large\hy{}scale connectomic researches\cite{ref21,ref22,ref23,ref24,ref25,ref26}. A flood\hy{}filling network\hy{}based architecture\cite{ref20,ref22} has been implemented on the human brain tissue vEM dataset to segment neurons and other cellular components, including astrocytes, in the human temporal cortex. A more recent 3D EM segmentation approach has further improved the efficiency of dense connectomic reconstruction through strategies such as local shape descriptor\hy{}based affinity learning and scalable block\hy{}wise processing\cite{ref27}. Since this approach was developed for general connectomics, it prioritizes large\hy{}scale neural circuit reconstruction rather than astrocyte\hy{}centric reconstruction. Although such general\hy{}purpose approaches may also be applicable to glial structures, they are not specifically optimized for preserving the irregular, and highly branched morphology of astrocytic processes or for astrocyte\hy{}focused morphometric workflows. This leaves a methodological gap for computationally efficient, astrocyte\hy{}centric segmentation and reconstruction tailored to anisotropic vEM data and downstream quantitative morphological analysis. In addition, current 3D approaches are often constrained by high computational demands, reduced robustness to anisotropic vEM characterized by high in\hy{}plane resolution but substantially coarser axial sampling, and difficulty in directly processing high\hy{}resolution inputs. They further require large\hy{}scale annotated volumetric datasets, whose acquisition is costly and labor\hy{}intensive. To overcome these limitations, two\hy{}dimensional (2D) segmentation architectures provide a practical and lightweight alternative for astrocyte vEM, balancing reconstruction fidelity with computational efficiency. By leveraging rich planar details while avoiding the heavy memory and computational demands associated with volumetric 3D models, 2D models consume less GPU memory, facilitate faster inference, and can be trained effectively on smaller annotated datasets.

\begin{figure*}[tbp]
\centering
\includegraphics[width=\textwidth]{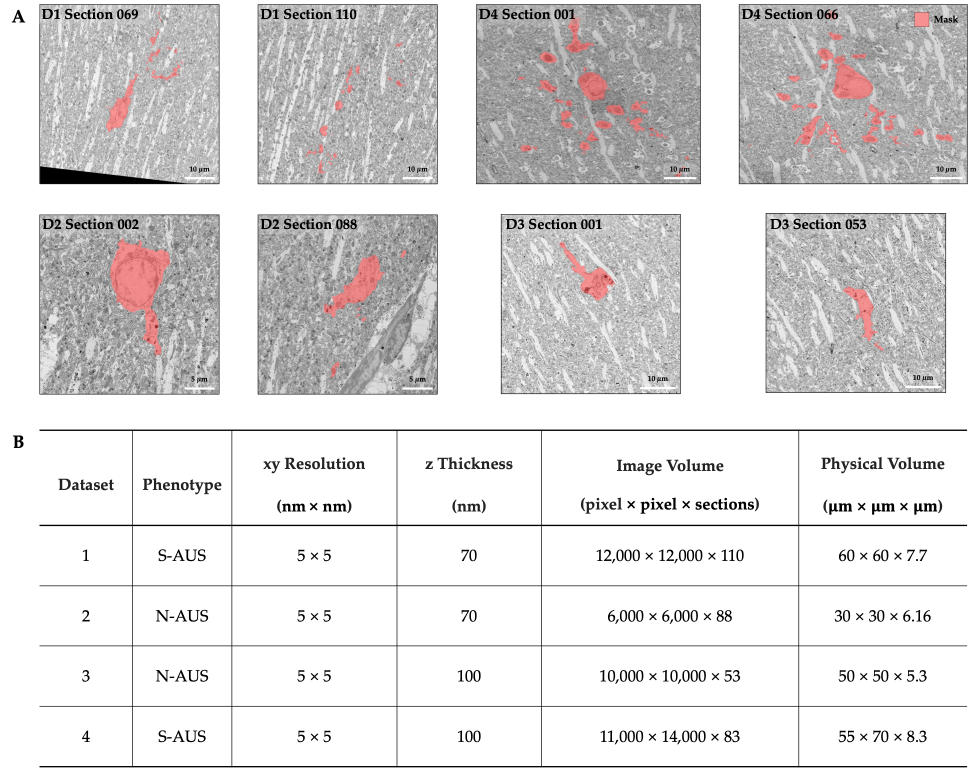}
\caption{\textbf{Visualization and statistics of astrocyte morphological profiles across vEM datasets.} \textbf{A} Representative EM sections from astrocytes under stress (S-AUS) datasets (D1, D4) featuring complex, hyper\hy{}ramified morphologies and normal (N-AUS) datasets (D2, D3) exhibiting standard baseline morphologies. Red overlays indicate astrocyte masks. \textbf{B} Summary of dataset statistics and imaging parameters across four vEM datasets. Panel A overlays are downsampled for display from the native 5 $\times$ 5 nm$^{2}$ in\hy{}plane acquisition resolution to effective in\hy{}plane pixel (px) spacings of 58.6 nm/px (D1), 29.3 nm/px (D2), 48.8 nm/px (D3), and 68.4 nm/px (D4).}
\label{fig:datasets}
\end{figure*}

In practice, this 2D modeling paradigm is commonly implemented using convolutional neural network (CNN) architectures such as U-Net\cite{ref28} and its widely adopted 2D variants\cite{ref29,ref30,ref31,ref32}, as well as context\hy{}aggregation models including DeepLab\cite{ref33,ref34,ref35,ref36}, which have demonstrated strong performance in 2D image segmentation. More recently, transformer\hy{}based\cite{ref37,ref38,ref39} and hybrid CNN--transformer architectures\cite{ref40,ref41} have further improved the modeling of long\hy{}range spatial dependencies. Nevertheless, their limited input sizes, typically a few hundred pixels per side, restrict their applicability to high\hy{}resolution EM images. As a practical solution, patch\hy{}based 2D segmentation pipelines are widely adopted in EM image analysis\cite{ref42,ref43,ref44,ref45,ref46}, enabling scalable training and inference on high\hy{}resolution EM sections by subdividing them into manageable patches while preserving local ultrastructural details. Yet the design and impact of patch sampling strategies remain underexplored. This limitation is particularly pronounced in astrocyte vEM segmentation, where astrocyte\hy{}positive regions (foreground) occupy only a small and spatially fragmented fraction of each section due to the characteristic star\hy{}shaped morphology of astrocytes (Fig.~\ref{fig:datasets}A). The complex geometry of astrocytes makes accurate segmentation and subsequent volumetric reconstruction from vEM data challenging\cite{ref8,ref9}. Because section\hy{}wise 2D segmentation does not explicitly model inter\hy{}section continuity, morphology\hy{}preserving post\hy{}processing is used as a lightweight 3D\hy{}aware step to compensate for this missing volumetric context and to assemble consistent 3D astrocyte structures from 2D predictions. Compared with native 3D model training and inference, this post\hy{}processing step uses a small number of interpretable parameters, runs without GPU acceleration, and adds relatively low computational overhead while improving volumetric consistency.

These limitations are particularly critical in the hippocampal stratum radiatum, a region densely populated with CA1 apical dendrites and excitatory synapses, where astrocyte processes are positioned to regulate synaptic transmission, extracellular diffusion, and activity\hy{}dependent plasticity\cite{ref5,ref47}. EM enables direct visualization of perisynaptic astrocytic processes and their physical relationship with synapses, providing a more accurate and disease\hy{}relevant assessment of astrocyte morphology and its potential impact on circuit function\cite{ref47}.

\begin{figure*}[tbp]
\centering
\includegraphics[height=0.66\textheight,keepaspectratio]{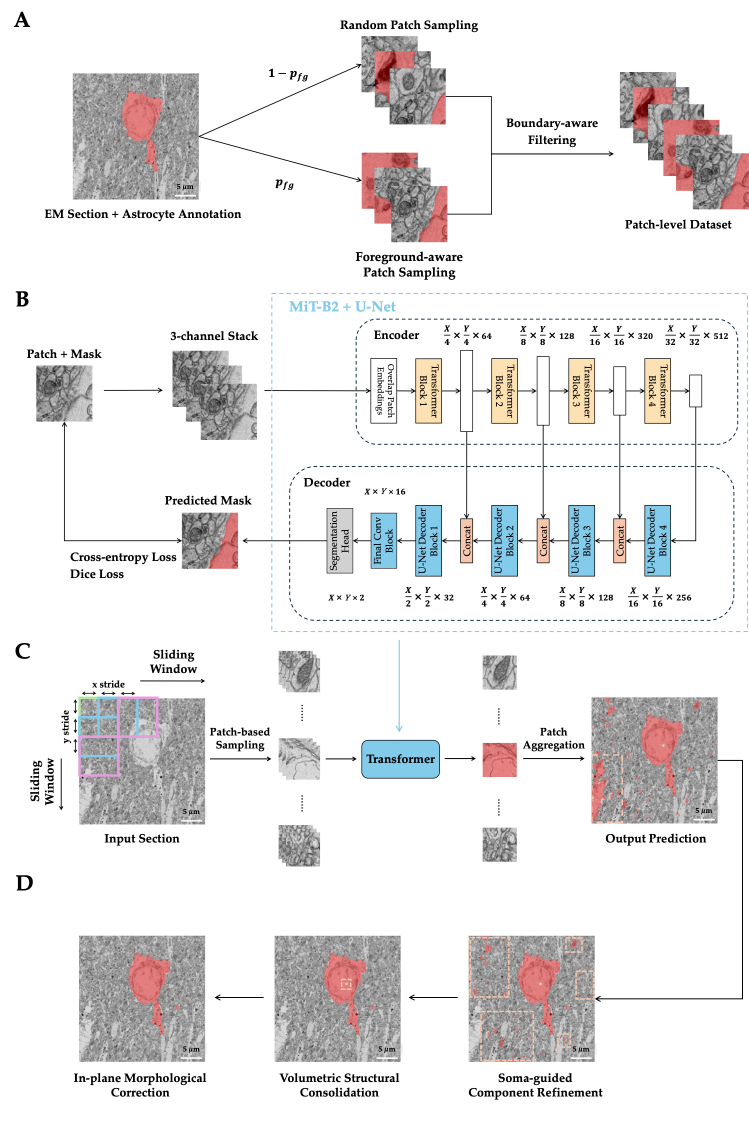}
\caption{\textbf{Lightweight 2D astrocyte segmentation and morphology\hy{}preserving reconstruction in AstroSeg-vEM.} \textbf{A} Patch\hy{}level dataset generation leveraging foreground\hy{}aware sampling. \textbf{B} Hierarchical transformer\hy{}based U-Net with a MiT-B2 encoder capturing multi\hy{}scale morphological features through self\hy{}attention. \textbf{C} Overlapping sliding\hy{}window inference strategy ensuring in\hy{}plane spatial continuity via probabilistic aggregation (See Methods). \textbf{D} Morphology\hy{}preserving post\hy{}processing by sequentially performing soma\hy{}guided component refinement, volumetric structural consolidation, and in\hy{}plane morphological correction to filter spatially inconsistent artifacts and refine the connectivity of astrocytic processes across vEM sections. Full\hy{}section overviews and post\hy{}processing renders in A, C, and D are downsampled for display to an effective in\hy{}plane pixel (px) spacing of 29.3 nm/px.}
\label{fig:pipeline}
\end{figure*}

Here, we propose AstroSeg-vEM, a lightweight, astrocyte\hy{}centric deep learning framework for astrocyte labeling and analysis. AstroSeg-vEM integrates automated 2D segmentation with morphology\hy{}preserving reconstruction for large\hy{}scale astrocyte analysis in vEM data (Fig.~\ref{fig:pipeline}). To our knowledge, this represents one of the first demonstrations of an efficient 2D deep learning framework for automated astrocyte segmentation and morphology\hy{}preserving 3D reconstruction, with outputs designed to support downstream expert\hy{}guided proofreading and analysis. The framework operates directly on high\hy{}resolution EM sections and is optimized for the non\hy{}regional, highly interconnected morphology of astrocytes, balancing between reconstruction fidelity and computational efficiency. Crucially, this unified framework not only enables quantitative characterization of astrocytic ultrastructure states but also provides an intrinsic metric for evaluating segmentation performance.

\section{Results}

\subsection{Astrocyte datasets and experimental setup}

We evaluate the robustness of AstroSeg-vEM across different ultrastructural conditions using four independent vEM datasets (D1, D2, D3, and D4 in Fig.~\ref{fig:datasets}). All datasets were acquired from the mouse hippocampus by array tomography, and key acquisition parameters, including strongly anisotropic voxel spacing, image and physical volumes, are summarized in Fig.~\ref{fig:datasets}B. Note that the datasets exhibit pronounced voxel anisotropy with axial thicknesses (70 or 100 nm) that are 14 to 20 times larger than the in\hy{}plane resolution (5 $\times$ 5 nm$^{2}$). They capture expansive in\hy{}plane fields of view of up to 60 $\times$ 60 \textmu{}m$^{2}$ (D1) and 55 $\times$ 70 \textmu{}m$^{2}$ (D4), while their axial coverage is limited by the finite number of sections. Based on these characteristics, the datasets encompass two distinct astrocyte profiles: normal and stress\hy{}associated astrocyte ultrastructural states (Fig.~\ref{fig:datasets}A). Here, astrocyte ultrastructural states (AUS) refer to distinct electron microscopy\hy{}resolvable three\hy{}dimensional configurations of astrocyte morphology, including process architecture under normal (N-AUS) and stress\hy{}associated (S-AUS) conditions.

Three distinct experimental settings (Table~\ref{tab:design}) are defined to systematically evaluate the robustness of 2D segmentation and post\hy{}processing pipeline in AstroSeg-vEM across N-AUS and S-AUS. Experiments 1 (Exp 1) and 2 (Exp 2) establish a morphological cross\hy{}validation by training on S-AUS datasets and testing on N-AUS datasets and vice versa. Experiment 3 (Exp 3) assesses the full AstroSeg-vEM framework for biological analysis, training on three datasets (D1, D2, D3) and reserving D4 as a held\hy{}out test dataset for 3D reconstruction and quantitative morphological analysis.

\begin{table*}[tbp]
\caption{\textbf{Experimental design and dataset allocation}}
\label{tab:design}
\begin{ruledtabular}
\begin{tabular}{llllll}
\textbf{Experiment} & \textbf{Objective} & \textbf{Training Set} & \textbf{Validation Set} & \textbf{Testing Set} & \textbf{Downstream Tasks} \\
\colrule\noalign{\vskip 3pt}
Exp 1 & \stk{Cross\hy{}domain}{(S-AUS $\rightarrow$ N-AUS)} & \stk{D1, D4}{(first 80\%)} & \stk{D1, D4}{(last 20\%)} & D2, D3 & Segmentation Accuracy \\[8pt]
Exp 2 & \stk{Cross\hy{}domain}{(N-AUS $\rightarrow$ S-AUS)} & \stk{D2, D3}{(first 80\%)} & \stk{D2, D3}{(last 20\%)} & D1, D4 & Segmentation Accuracy \\[8pt]
Exp 3 & \stk{Full Pipeline}{\& Morphology} & \stk{D1, D2, D3}{(first 80\%)} & \stk{D1, D2, D3}{(last 20\%)} & D4 & 3D Reconstruction \& Analysis \\
\end{tabular}
\end{ruledtabular}
\end{table*}

\subsection{Astrocyte\hy{}centric segmentation across diverse ultrastructural states}

To validate the core capabilities of the transformer\hy{}based segmentation model within AstroSeg-vEM, we compared its raw, unprocessed segmentation performance with representative baselines, including U-Net\cite{ref28} and SegFormer\cite{ref39}, as well as Swin-Unet\cite{ref37}, under the same experimental setting (Exp 3). Segmentation performance was evaluated using four metrics: Dice score (Dice), weighted Dice (wDice), precision, and recall; the mathematical definition of wDice is written in Methods. Table~\ref{tab:arch} summarizes that the proposed model achieves higher Dice and precision, while maintaining comparable performance in wDice and recall. It attains the highest Dice score of 0.70 $\pm$ 0.02 and precision of 0.58 $\pm$ 0.03, outperforming both SegFormer and Swin-Unet, with a modest but consistent paired advantage over SegFormer across sections (Fig. S3), whereas U-Net exhibits markedly degraded performance (Dice = 0.36 $\pm$ 0.05). Consistently high performance of AstroSeg-vEM with wDice (0.94 $\pm$ 0.02) and recall (0.89 $\pm$ 0.04) indicates that primary astrocyte structures are consistently captured. Improvements in Dice and precision reflect enhanced delineation of fragmented astrocytic processes, highlighting the advantage of transformer\hy{}based architectures in modeling fine\hy{}scale spatial dependencies. Despite superior segmentation performance, our AstroSeg-vEM introduces only a small increase in model size (+11.7\%) and computational cost (+28.7\% GFLOPs) relative to SegFormer, while achieving 7.5\% faster inference time per section. Compared with U-Net, it reduces computational cost by 89.2\% and inference time by 50.8\% with 20.1\% fewer parameters. Relative to Swin-Unet, AstroSeg-vEM further reduces parameter count by 33.3\%, computational cost by 39.4\%, and inference time by 14.0\%. These results demonstrate that AstroSeg-vEM optimizes the trade\hy{}off between segmentation accuracy, runtime efficiency, and model complexity, making it well\hy{}suited for capturing the complex and highly ramified astrocyte morphology.

\begin{table*}[tbp]
\caption{\textbf{Segmentation Architectures Performance and Efficiency Comparison} Dice, wDice, precision, and recall are mean $\pm$ standard deviation across \emph{n} = 83 sections from a single held\hy{}out volume (Exp 3 D4), not independent biological replicates. Values are computed at the native 5 nm resolution and are not directly comparable with the 20 nm common\hy{}grid values of the 3D nnU-Net comparison (see Discussion).}
\label{tab:arch}
\begin{ruledtabular}
\begin{tabular}{lccccccc}
\textbf{Architecture} & \textbf{Dice} & \textbf{wDice} & \textbf{Precision} & \textbf{Recall} & \textbf{Inference Time} & \textbf{Parameters} & \textbf{GFLOPs} \\
& \textbf{(Exp 3)} & \textbf{(Exp 3)} & \textbf{(Exp 3)} & \textbf{(Exp 3)} & \textbf{per Section (s)} & & \\
\colrule\noalign{\vskip 3pt}
AstroSeg-vEM (Ours) & {\bfseries\boldmath 0.70 $\pm$ 0.02} & {\bfseries\boldmath 0.94 $\pm$ 0.02} & {\bfseries\boldmath 0.58 $\pm$ 0.03} & {\bfseries\boldmath 0.89 $\pm$ 0.04} & \textbf{101.96} & 27.60M & 10.80 \\
SegFormer\cite{ref39} & 0.68 $\pm$ 0.02 & 0.94 $\pm$ 0.03 & 0.55 $\pm$ 0.03 & {\bfseries\boldmath 0.89 $\pm$ 0.04} & 110.23 & \textbf{24.72M} & \textbf{8.39} \\
U-Net\cite{ref28} & 0.36 $\pm$ 0.05 & 0.91 $\pm$ 0.02 & 0.23 $\pm$ 0.04 & 0.87 $\pm$ 0.04 & 207.37 & 34.53M & 100.07 \\
Swin-Unet\cite{ref37} & 0.65 $\pm$ 0.06 & 0.90 $\pm$ 0.10 & 0.54 $\pm$ 0.03 & 0.84 $\pm$ 0.12 & 118.53 & 41.38M & 17.82 \\
\end{tabular}
\end{ruledtabular}
\end{table*}

\begin{figure*}[tbp]
\centering
\includegraphics[height=0.68\textheight,keepaspectratio]{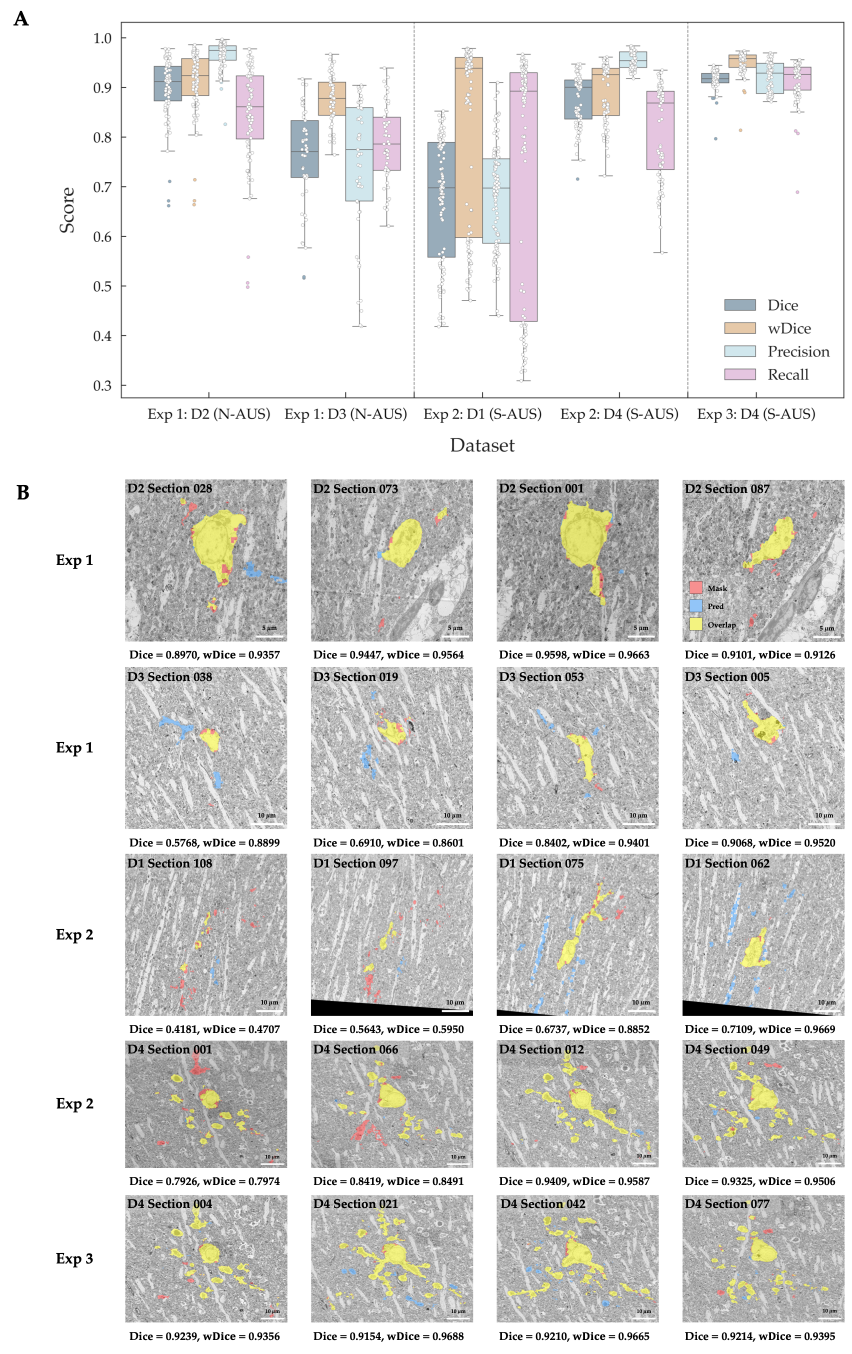}
\caption{\textbf{Quantitative robustness and pixel\hy{}level morphological fidelity of the astrocyte segmentation pipeline.} \textbf{A} Quantitative segmentation performance. Boxplots (median and interquartile ranges) and individual section scores (scatter) summarize metrics across three experiments. \textbf{B} Segmentation visual comparison. Representative pixel\hy{}level results for selected sections are collected across Exp 1--3. Yellow = true positives (overlap), red = false negatives (missed), and blue = false positives (over\hy{}segmentation). Sections span a range of Dice and wDice scores and morphological regions, not only the highest\hy{}scoring examples. Panel B overlays are downsampled for display from the native 5 $\times$ 5 nm$^{2}$ in\hy{}plane acquisition resolution to effective in\hy{}plane pixel (px) spacings of 58.6 nm/px (D1), 29.3 nm/px (D2), 48.8 nm/px (D3), and 68.4 nm/px (D4).}
\label{fig:performance}
\end{figure*}

Building upon robust raw predictions, we evaluated the full segmentation pipeline including a dedicated post\hy{}processing module. The distributions of Dice, wDice, precision, and recall across Exp 1--3 are summarized in Fig.~\ref{fig:performance}A and Table~\ref{tab:robustness}. Robustness was assessed through morphological cross\hy{}validation (Exp 1 and Exp 2). Within this cross\hy{}validation phase, N-AUS datasets (D2, D3) generally exhibit more compact interquartile ranges of four metrics compared to the S-AUS datasets (D1, D4). The increased variance and extended lower\hy{}scoring tails in the S-AUS datasets are concentrated in a subset of sections where the ground\hy{}truth astrocytic processes are sparse and thin, whereas the post\hy{}processing threshold is mainly tuned for large and coherent astrocyte regions (Fig. S1). Across datasets, Dice and recall generally show broader interquartile ranges and lower medians than precision and wDice. Nevertheless, performance remains sufficient for large\hy{}scale morphological discovery, as the median Dice and recall values are consistently above 0.69 and 0.78, respectively. Notably, this cross\hy{}validation utilizes a uniform set of hyperparameters, demonstrating that our method effectively generalizes across topological features without individual tuning for each dataset. A detailed discussion of why proposed wDice is a good complementary metric for astrocyte segmentation is provided in the Discussion.

Following the cross\hy{}validation, we trained a more comprehensive model on combined N-AUS and S-AUS datasets to evaluate it on the held\hy{}out D4 (Exp 3). Recall remains stable with a median of 0.93, while wDice reaches its peak median of 0.96. For Exp 3, the post\hy{}processing parameters are further optimized on D1--D3, without using D4, to maximize the structural continuity necessary for high\hy{}fidelity 3D reconstruction. By bridging the gap between raw pixel identification and volumetric connectivity, the pipeline provides cleaner section\hy{}wise predictions for high\hy{}fidelity 3D reconstruction and downstream proofreading. In order to test how the amount of training data affects performance, we additionally evaluated two protocols with fewer training volumes, two\hy{}fold cross\hy{}validation (2-fold) and leave\hy{}one\hy{}volume\hy{}out (LOVO) cross\hy{}validation. Their training, validation and test volumes are listed in Supplementary Table S1. Exp 3 was trained on D1, D2 and D3, with the last 20\% of sections of each volume held out for validation, whereas LOVO fold 1 was trained on D1 and D2 and validated on D3, so the two protocols differ in the amount of training data as well as in validation strategy. On the 83 sections of D4, post\hy{}processed Dice decreased from 0.92 $\pm$ 0.02 in Exp 3 to 0.71 $\pm$ 0.04 in LOVO fold 1 (Supplementary Table S1), and raw Dice decreased from 0.70 $\pm$ 0.02 to 0.58 $\pm$ 0.04, which we interpret as limited robustness when fewer training volumes are available. In 2-fold-1, which was trained on D1 and validated on D4, post\hy{}processed Dice and wDice on the 88 sections of D2 were 0.42 $\pm$ 0.43 and 0.47 $\pm$ 0.43, compared with 0.53 $\pm$ 0.18 and 0.94 $\pm$ 0.05 before post\hy{}processing. D4 served as a test\hy{}only volume in Exp 3 and LOVO.

AstroSeg-vEM incorporates three post\hy{}processing modules designed to refine segmentation outputs and improve structural coherence: soma\hy{}guided component refinement (SCR), volumetric structural consolidation (VSC), and in\hy{}plane morphological correction (IMC). A systematic ablation study (Table~\ref{tab:ablation}) shows that the AstroSeg-vEM post\hy{}processing pipeline consistently improves Dice in all five evaluated settings, but not always wDice. Already high at baseline, wDice is unchanged or slightly higher in Exp 1 and Exp 3 and decreases in Exp 2, from 0.91 to 0.82 in D1 and from 0.94 to 0.89 in D4. Although the baseline segmentation model already achieves high wDice (0.87--0.94), indicating that primary astrocyte structures are largely captured, Dice remains low due to false predictions. The SCR module reduces spatial inconsistencies, leading to higher Dice, especially in Exp 1 D2 and Exp 2 D1. The VSC module provides the largest gain by enforcing global 3D connectivity, raising Dice from 0.72 to 0.88 in Exp 1 D2 and from 0.43 to 0.67 in Exp 2 D1. The IMC module further refines topology, yielding smaller but consistent gains (e.g. 0.88 to 0.90 in Exp 1 D2). Note that wDice remains relatively stable throughout the ablation process in Exp 1 and Exp 3, suggesting that core astrocytic structures are already identified by the segmentation model, whereas post\hy{}processing primarily restores structural continuity rather than improving pixel\hy{}wise overlap. These results highlight the importance of morphology\hy{}preserving post\hy{}processing for producing structurally coherent representations suitable for 3D reconstruction and expert\hy{}guided proofreading. Consistent with this, a per\hy{}section decomposition of residual errors into the false discovery rate (FDR) and the miss rate (Methods) shows that post\hy{}processing lowers the FDR markedly in all five settings (Fig. S2A), whereas the miss rate is similar or lower in Exp 1 and Exp 3 and higher in Exp 2 (Fig. S2B). Because the two rates have different denominators, predicted\hy{}positive pixels for the FDR and ground\hy{}truth\hy{}positive pixels for the miss rate, they are presented as complementary error measures. Component\hy{}level analysis of the Exp 3 D4 volume gives the same picture (Supplementary Table S2). Post\hy{}processing dramatically reduced the mean number of predicted connected components per section by approximately 90\%, from 655.7 to 66.3 (ground truth, 29.1), and the mean number of spurious components per section by approximately 95\%, from 599.9 to 30.3. The mean number of split ground\hy{}truth components per section also decreased from 10.0 to 5.0, whereas missed ground\hy{}truth components increased from 0.9 to 3.4 and merges were rare and unchanged (0.5 per section).

\begin{table*}[tbp]
\caption{\textbf{Ablation study of the post\hy{}processing pipeline for astrocyte segmentation} Dice and wDice are mean $\pm$ standard deviation across sections within each held\hy{}out dataset (Exp 1 D2, \emph{n} = 88; Exp 1 D3, \emph{n} = 53; Exp 2 D1, \emph{n} = 110; Exp 2 D4, \emph{n} = 83; Exp 3 D4, \emph{n} = 83), not independent biological replicates. All values are computed at the native 5 nm resolution and are not directly comparable with the 20 nm common\hy{}grid values of the 3D nnU-Net comparison (see Discussion).}
\label{tab:ablation}
\begin{ruledtabular}
\setlength{\tabcolsep}{4pt}
\renewcommand{\arraystretch}{1.15}
\begin{tabular}{ccccccccccccc}
\multicolumn{3}{c}{\textbf{Post\hy{}processing}} & \multicolumn{2}{c}{\textbf{Exp 1 D2}} & \multicolumn{2}{c}{\textbf{Exp 1 D3}} & \multicolumn{2}{c}{\textbf{Exp 2 D1}} & \multicolumn{2}{c}{\textbf{Exp 2 D4}} & \multicolumn{2}{c}{\textbf{Exp 3 D4}} \\
\textbf{SCR} & \textbf{VSC} & \textbf{IMC} & \textbf{Dice} & \textbf{wDice} & \textbf{Dice} & \textbf{wDice} & \textbf{Dice} & \textbf{wDice} & \textbf{Dice} & \textbf{wDice} & \textbf{Dice} & \textbf{wDice} \\
\colrule\noalign{\vskip 3pt}
 & & & \ms{0.57}{0.19} & {\boldmath\ms{\textbf{0.91}}{\textbf{0.06}}} & \ms{0.50}{0.13} & \ms{0.87}{0.05} & \ms{0.38}{0.09} & {\boldmath\ms{\textbf{0.91}}{\textbf{0.05}}} & \ms{0.71}{0.02} & {\boldmath\ms{\textbf{0.94}}{\textbf{0.02}}} & \ms{0.70}{0.02} & \ms{0.94}{0.02} \\
$\checkmark$ & & & \ms{0.72}{0.09} & \ms{0.90}{0.07} & \ms{0.52}{0.12} & \ms{0.87}{0.05} & \ms{0.43}{0.10} & \ms{0.89}{0.08} & \ms{0.71}{0.03} & \ms{0.92}{0.03} & \ms{0.70}{0.02} & \ms{0.93}{0.02} \\
$\checkmark$ & $\checkmark$ & & \ms{0.88}{0.07} & \ms{0.90}{0.07} & \ms{0.75}{0.10} & \ms{0.86}{0.05} & \ms{0.67}{0.12} & \ms{0.81}{0.18} & \ms{0.85}{0.05} & \ms{0.87}{0.06} & \ms{0.89}{0.02} & \ms{0.93}{0.03} \\
$\checkmark$ & $\checkmark$ & $\checkmark$ & {\boldmath\ms{\textbf{0.90}}{\textbf{0.06}}} & {\boldmath\ms{\textbf{0.91}}{\textbf{0.06}}} & {\boldmath\ms{\textbf{0.76}}{\textbf{0.10}}} & {\boldmath\ms{\textbf{0.88}}{\textbf{0.05}}} & {\boldmath\ms{\textbf{0.68}}{\textbf{0.12}}} & \ms{0.82}{0.18} & {\boldmath\ms{\textbf{0.87}}{\textbf{0.05}}} & \ms{0.89}{0.06} & {\boldmath\ms{\textbf{0.92}}{\textbf{0.02}}} & {\boldmath\ms{\textbf{0.95}}{\textbf{0.02}}} \\
\end{tabular}
\end{ruledtabular}
\end{table*}

The S-AUS datasets pose extreme challenges for automated 2D segmentation due to fragmented membrane signals and atypical morphological distortions, whose biological complexities often lead to severe topological discontinuities in standard frameworks. As shown in the pixel\hy{}level comparison (Fig.~\ref{fig:performance}B), our pipeline demonstrates remarkable robustness across representative z sections. Sections were selected by jointly considering quantitative performance (Dice and wDice) and morphological diversity spanning different score ranges and astrocytic structures across both N-AUS and S-AUS datasets rather than only the highest\hy{}scoring examples. Predominant yellow regions confirm a precise pixel\hy{}wise overlap that is particularly striking in the proximal regions, where the soma and primary branches are recovered with high fidelity. This high fidelity in the morphological center is critical, as these structures define the spatial territory and basic scaffold of the astrocyte.

While minor misclassifications (indicated by red and blue pixels) are occasionally detectable in the distal processes far from the astrocyte center, these are primarily restricted to the peripheral fringes where EM signals approach the limits of resolvable contrast. Importantly, even in these challenging distal zones, the pipeline maintains a high degree of directional consistency and reduces merge errors that can compromise downstream reconstruction\cite{ref48}. This high precision in the primary and secondary branching structures provides the necessary confidence to proceed with the full 3D reconstruction of the 2D stacks. By preserving the integrity of the main morphological scaffold, our pipeline enables volumetric representations that reflect the underlying biological structure rather than reconstruction artifacts.

\subsection{3D reconstruction and comparative morphological analysis}

\begin{figure}[tbp]
\centering
\includegraphics[width=0.9\columnwidth]{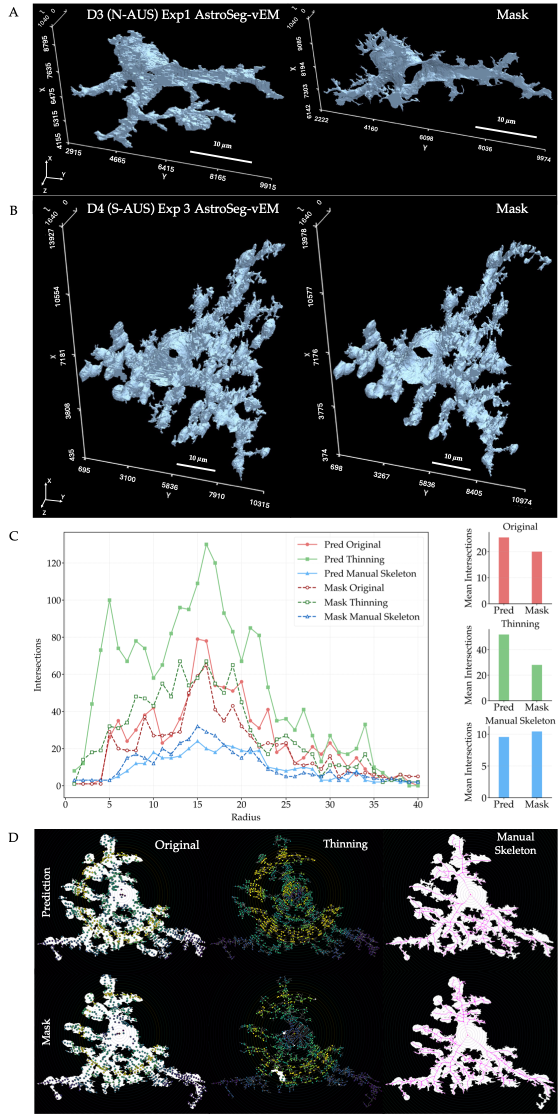}
\caption{\textbf{Morphology\hy{}preserving 3D reconstruction (downsampled ten times and anisotropically stretched back to the original voxel scale) and quantitative analysis of astrocytes using AstroSeg-vEM.} AstroSeg-vEM reconstruction (left) and the ground\hy{}truth masks (right) of N-AUS D3 from Exp 1 (\textbf{A}) and S-AUS D4 from Exp 3 (\textbf{B}). \textbf{C} Sholl plots (left) and histograms of mean intersections (right) between manual masks and AstroSeg-vEM\hy{}predicted masks of S-AUS D4 from Exp 3 for original, thinning, and manual methods. \textbf{D} Visualization of Sholl analysis results with AstroSeg-vEM (top) and manual (bottom) masks for three Fiji SNT plugin approaches. Voxel spacing after stretching back to native scale: 5 $\times$ 5 $\times$ 100 nm$^{3}$ (D3 in A, D4 in B).}
\label{fig:reconstruction}
\end{figure}

To visually assess the reconstruction quality, we generated a 3D surface mesh from the ten times downsampled post\hy{}processed segmented vEM stack in Exp 3 (Fig.~\ref{fig:reconstruction}B) and anisotropically stretched it back to the original voxel scale. The AstroSeg-vEM\hy{}derived mesh successfully extracts the complex, hyper\hy{}ramified morphology of astrocytes, demonstrating high visual fidelity and structural congruence with the ground\hy{}truth manual annotations across both somatic regions and complex distal processes. Sholl analysis is commonly used to quantify astrocyte morphological complexity\cite{ref11,ref49,ref50,ref51,ref52,ref53}. For evaluating whether our predicted labels preserve astrocyte morphology for downstream analysis, we performed Sholl analysis (Figs.~\ref{fig:reconstruction}C and \ref{fig:reconstruction}D). Intersection counts were computed using three approaches with the Fiji SNT plugin\cite{ref54}: Direct analysis of segmented images (original), analysis after auto\hy{}skeletonization (thinning), and analysis after manual skeletonization (manual) in Figs.~\ref{fig:reconstruction}C and \ref{fig:reconstruction}D. Auto\hy{}skeletonization resulted in an overestimation of mean intersections and a shift in the Sholl profile along the radial axis, likely due to spurious branches introduced by simple thinning\hy{}based algorithms. In contrast, direct mask\hy{}based analysis and manual skeletonization produced similar mean intersections and comparable Sholl plot profiles. These results indicate that the AstroSeg-vEM framework preserves astrocyte morphology and is suitable for downstream morphometric analysis.

\section{Discussion}

In this study, we develop AstroSeg-vEM as a lightweight, astrocyte\hy{}centric vEM reconstruction pipeline that combines transformer\hy{}based 2D segmentation with morphology\hy{}preserving post\hy{}processing. We position AstroSeg-vEM not as evidence of a general superiority of 2D over 3D segmentation, but as a substantially lighter, more memory\hy{}efficient alternative suited to accelerating native\hy{}resolution, manual vEM annotation workflows. The approach demonstrates excellent performance across different ultrastructure and morphological conditions, achieving high segmentation accuracy with efficient computation.

Astrocytes are increasingly recognized as structurally complex partners of synapses rather than passive supporting cells. In particular, perisynaptic astrocytic processes occupy nanometer\hy{}scale domains around excitatory synapses and regulate neurotransmitter clearance, ionic homeostasis, extracellular diffusion, and metabolic support\cite{ref5,ref12,ref47}. Classical serial EM studies in hippocampal CA1 demonstrated that astrocytic processes contact only a subset of synapses and often cover only part of the synaptic perimeter, indicating that astrocyte\hy{}synapse relationships are spatially selective rather than uniform\cite{ref5,ref6}. Therefore, the ultrastructural differences detected in the present study may reflect remodeling of the local synaptic microenvironment rather than global astrocyte hypertrophy alone. By enabling 3D reconstruction of astrocytic processes under normal and stress\hy{}associated conditions, AstroSeg-vEM provides a practical framework for quantifying astrocyte process architecture, synaptic apposition, and spatial occupancy at a resolution that cannot be achieved by light microscopy.

Note that S-AUS should not be interpreted as synonymous with molecularly defined reactive astrogliosis or as direct evidence of causal remodeling. Instead, we operationally define structural remodeling as nanoscale reorganization of astrocytic processes within the synaptic microenvironment, including changes in perisynaptic process geometry, synaptic apposition, and spatial occupancy. These features are not readily accessible using light microscopy\cite{ref5,ref11}, but can be evaluated by vEM\hy{}based reconstruction. Accordingly, our results describe vEM\hy{}defined structural states characterized by altered astrocytic process organization, providing a foundation for future studies to investigate how these structural changes influence spatial relationships with neighboring dendrites. Future studies integrating vEM with molecular markers of astrocyte identity, reactivity, metabolic stress, and inflammatory signaling will be required to determine these ultrastructural states in terms of transcriptomic or immunohistochemical astrocyte subtypes.

For better structural fidelity assessment, we introduced wDice as a complementary metric whose formal definition is provided in the Methods. By placing greater emphasis on local structural agreement while reducing the influence of spatially distant artifacts, wDice is particularly informative for evaluating preservation of the main morphological structures of astrocytes. According to the ablation study (Table~\ref{tab:ablation}), wDice remains high (0.81--0.95) across all experimental stages and changes by at most 0.02 in Exp 1 and Exp 3, whereas it decreases in Exp 2, indicating consistent preservation of astrocytic structure, even in branches with peripheral noise. The biological relevance of wDice is further supported by the 3D reconstruction and Sholl analysis results (Fig.~\ref{fig:reconstruction}), which show that the predicted segmentations preserve the overall morphological profiles of manual annotations.

For possible condition batch confounding, all samples were processed leveraging the same solutions and processing time to minimize preparation\hy{}related variability. Although section thickness differed between datasets (70 or 100 nm), comparison of one control and one stress dataset at each thickness did not reveal an apparent thickness\hy{}dependent effect on segmentation performance, and we therefore consider section thickness unlikely to be the main source of between\hy{}dataset differences. Nevertheless, biological variability among animals and technical variability related to sectioning and imaging cannot be completely excluded. D1 and D2 were prepared on the same day, as were D3 and D4, and the four datasets were therefore prepared in two separate batches. D3 and D4 were imaged in the same session, whereas D1 and D2 were each imaged in a separate session. No apparent batch- or imaging\hy{}session\hy{}level effect was observed. Because an independent inter\hy{}annotator agreement analysis could not be performed retrospectively for the full dataset, the ground\hy{}truth annotations may contain annotator\hy{}specific uncertainty, and annotator and dataset effects cannot be completely separated in the current dataset. Finally, cross\hy{}volume robustness is limited when fewer training volumes are available. Post\hy{}processed Dice on D4 decreased from 0.92 $\pm$ 0.02 with three training volumes (Exp 3) to 0.71 $\pm$ 0.04 with two (LOVO fold 1), and all experiments rest on four volumes from four animals.

The AstroSeg-vEM framework also demonstrates high computational efficiency. Across Exp 1 and Exp 2, the per\hy{}section inference time ranged from 23.63 s to 102.47 s depending on dataset size and image dimensions (Table~\ref{tab:runtime}). In Exp 3, our transformer\hy{}based model maintains superior segmentation performance while achieving substantially lower inference time than U-Net and Swin-Unet on D4 (Table~\ref{tab:arch}). This high throughput is achieved on consumer\hy{}grade hardware (NVIDIA RTX 2060, 12 GB VRAM), and the morphology\hy{}preserving post\hy{}processing adds minimal total overhead (Table~\ref{tab:postruntime}). By bypassing the prohibitive memory requirements of native 3D CNNs, AstroSeg-vEM efficiently leverages the rich in\hy{}plane resolution of 2D sections to recover 3D structures.

\begin{table}[tbp]
\caption{\textbf{Segmentation Runtime Performance on Exp 1 and Exp 2 Leveraging Transformer\hy{}based model} Note: `Inference Time per Section' is the wall\hy{}clock time for loading, inferring and saving a single section and `Total Inference Time' is the wall\hy{}clock time of the whole evaluation run and additionally includes disk I/O and data\hy{}transfer overhead that is not attributed to individual sections.}
\label{tab:runtime}
\begin{ruledtabular}
\setlength{\tabcolsep}{4pt}
\begin{tabular}{lccc}
\textbf{Dataset} & \textbf{\cstk{Number of}{Sections}} & \textbf{\cstk{Inference Time}{\cstk{per Section}{(s)}}} & \textbf{\cstk{Total}{\cstk{Inference}{Time (s)}}} \\
\colrule\noalign{\vskip 3pt}
D1 S-AUS & 110 & 96.18 & 12598.69 \\
D2 N-AUS & 88 & 23.63 & 2443.60 \\
D3 N-AUS & 53 & 66.55 & 4152.01 \\
D4 S-AUS & 83 & 102.47 & 10148.54 \\
\end{tabular}
\end{ruledtabular}
\end{table}

\begin{table}[tbp]
\caption{\textbf{Morphology\hy{}preserving Post\hy{}processing Run Time Performance on Exp 1 and Exp 2}}
\label{tab:postruntime}
\begin{ruledtabular}
\setlength{\tabcolsep}{4pt}
\begin{tabular}{lcc}
\textbf{Dataset} & \textbf{\cstk{Number of}{Sections}} & \textbf{\cstk{Total Inference}{Time (s)}} \\
\colrule\noalign{\vskip 3pt}
D1 S-AUS & 110 & 520.81 \\
D2 N-AUS & 88 & 67.88 \\
D3 N-AUS & 53 & 111.47 \\
D4 S-AUS & 83 & 377.19 \\
\end{tabular}
\end{ruledtabular}
\end{table}

\begin{table*}[tbp]
\caption{\textbf{Quantitative robustness and pixel\hy{}level morphological fidelity of the astrocyte segmentation pipeline} Values are median and IQR across sections within each held\hy{}out dataset (Exp 1 D2, \emph{n} = 88; Exp 1 D3, \emph{n} = 53; Exp 2 D1, \emph{n} = 110; Exp 2 D4, \emph{n} = 83; Exp 3 D4, \emph{n} = 83), not independent biological replicates.}
\label{tab:robustness}
\begin{ruledtabular}
\begin{tabular}{lcccccccccc}
& \multicolumn{2}{c}{\textbf{Exp 1 D2}} & \multicolumn{2}{c}{\textbf{Exp 1 D3}} & \multicolumn{2}{c}{\textbf{Exp 2 D1}} & \multicolumn{2}{c}{\textbf{Exp 2 D4}} & \multicolumn{2}{c}{\textbf{Exp 3 D4}} \\
\textbf{Metric} & \textbf{Median} & \textbf{IQR} & \textbf{Median} & \textbf{IQR} & \textbf{Median} & \textbf{IQR} & \textbf{Median} & \textbf{IQR} & \textbf{Median} & \textbf{IQR} \\
\colrule\noalign{\vskip 3pt}
Dice      & 0.91 & 0.07 & 0.77 & 0.11 & 0.70 & 0.23 & 0.90 & 0.08 & 0.92 & 0.02 \\
wDice     & 0.92 & 0.07 & 0.88 & 0.07 & 0.94 & 0.36 & 0.93 & 0.10 & 0.96 & 0.02 \\
Precision & 0.97 & 0.03 & 0.77 & 0.19 & 0.70 & 0.17 & 0.95 & 0.03 & 0.93 & 0.06 \\
Recall    & 0.86 & 0.13 & 0.79 & 0.11 & 0.89 & 0.50 & 0.87 & 0.16 & 0.93 & 0.05 \\
\end{tabular}
\end{ruledtabular}
\end{table*}

At the backbone level, AstroSeg-vEM outperforms its closest competitor SegFormer by a mean paired Dice difference of only 0.021 across the 83 D4 sections (Table~\ref{tab:arch} and Fig. S3), which suggests that backbone performance is close to saturation for this task. This within\hy{}volume comparison is descriptive and makes no claim of inter\hy{}animal significance. The larger gain comes from the lightweight post\hy{}processing, which raises Dice on the same volume from 0.70 to 0.92 (Table~\ref{tab:ablation}), while the framework processes native 5 nm sections on a consumer\hy{}grade 12 GB GPU. The practical value of AstroSeg-vEM therefore lies in this post\hy{}processing stage and efficiency--accuracy trade\hy{}off rather than in a higher backbone accuracy ceiling.

A medium\hy{}weight, 3D block\hy{}wise system is a promising alternative. In a feasibility experiment, a 3D nnU-Net trained at 20 nm in\hy{}plane resolution with four times downsampled input, under the LOVO protocol with training on D1 and D2, reached a per\hy{}section Dice of 0.696 $\pm$ 0.055 on D4 without post\hy{}processing, compared with 0.916 $\pm$ 0.020 for the complete AstroSeg-vEM pipeline on the same 20 nm grid (Methods). Because the two models were not trained on the same volumes and nnU-Net received downsampled input and had no post\hy{}processing stage, we do not attribute this difference to the architecture. The 3D run required an NVIDIA A100 data\hy{}center GPU with 40 GB of memory and about 5.1 hours for 250 training epochs, and we did not attempt a 3D pipeline at native 5 nm resolution. We regard this as a limitation of the present study and a direction for future work.

Fok \emph{et al.}\cite{ref55} recently introduced a curvature\hy{}consistency deep learning approach that accelerates astrocyte reconstruction 12-fold and enables cross\hy{}species comparison of astrocyte nanoarchitecture. It was developed for near\hy{}isotropic FIB-SEM data (4 $\times$ 4 $\times$ 8 nm$^{3}$), whereas our array\hy{}tomography data are strongly anisotropic, with 5 nm pixels and 70--100 nm sections, which motivated our 2D design. Extending such 3D approaches to anisotropic data is a promising direction for future work.

While AstroSeg-vEM provides a high\hy{}fidelity foundation for mouse hippocampus astrocyte reconstruction, several aspects of the framework could be further improved. Further improvement in scalability for larger vEM volumes may be achieved by incorporating astrocyte\hy{}localization modules before high\hy{}resolution segmentation, thereby reducing computationally redundant processing of non\hy{}target tissue. Extending AstroSeg-vEM toward multi\hy{}class segmentation may also broaden its utility for studying astrocyte\hy{}synapse interactions in the hippocampal nano\hy{}environment. We envision that in future biological applications, analyses of process thickness distribution, synapse proximity, and organelle inclusion or contact\hy{}site quantification could further link astrocyte ultrastructural changes to functional alterations in neuron\hy{}glia communication.

\section{Methods}

\subsection{Sample preparation for vEM imaging}

Two controls and two with stress\hy{}associated astrocyte morphological changes identified by immunohistochemical analysis were used for electron microscopy (EM) analysis. These morphological changes were used to define the stress\hy{}associated condition and were not intended to indicate molecularly defined reactive astrogliosis. All animal experiments were conducted in accordance with the Institutional Animal Care and Use Committee (IACUC) guidelines and were approved by the IACUC under approval number IACUC-24-00067-M1. Mice were anesthetized with isoflurane and euthanized by cervical dislocation prior to tissue collection. They were subsequently perfused with 0.15 M cacodylate buffer (pH 7.4), followed by fixation with 2\% paraformaldehyde (PFA) and 2.5\% glutaraldehyde in the same buffer. The brains were removed and further fixed overnight at 4 \textdegree{}C in the same fixative. The fixed brains were then sectioned into 150-\textmu{}m\hy{}thick hippocampal slices using a vibratome (Leica). For scanning electron microscopy (SEM), tissue slices were post\hy{}fixed with 2\% osmium tetroxide (OsO$_{4}$; EMS) and 1.5\% potassium ferrocyanide (Sigma) in 0.15 M cacodylate buffer for 1 h on ice. The tissues were subsequently incubated in 1\% thiocarbohydrazide (TCH; TCI) for 20 min, followed by 2\% OsO$_{4}$ in distilled water for 30 min at room temperature. The samples were then stained with 2\% uranyl acetate (EMS) overnight at 4 \textdegree{}C, followed by incubation in lead aspartate solution for 30 min. The tissues were dehydrated through a graded ethanol series (20\%, 50\%, 70\%, 90\%, and 100\%) on ice. Following dehydration, the samples were sequentially infiltrated with 100\% acetone, mixtures of acetone and epoxy resin at ratios of 3:1, 1:1, and 1:3, and finally 100\% epoxy resin. The samples were then embedded using an EMbed-812 embedding kit (EMS) according to the manufacturer's instructions. For SEM imaging of the mouse hippocampus, serial thin sections (70 or 100 nm thick) were prepared from the resin\hy{}embedded samples using an ultramicrotome (EM UC7; Leica) and mounted on indium\hy{}tin\hy{}oxide\hy{}coated glass coverslips. Serial sections were imaged using a GeminiSEM 300 scanning electron microscope (Zeiss) equipped with Atlas 5 software at an accelerating voltage of 3.5 kV. Astrocytes in the stratum radiatum (SR) of the CA1 region were imaged at approximately 100--150 \textmu{}m from the CA1 pyramidal cell layer. Images were acquired at an x--y pixel size of 5 nm. D1--D4 were obtained from four independent mouse samples. D1 and D4 were tissue blocks from the hippocampus SR region exhibiting a stress\hy{}associated astrocyte phenotype, as identified by immunohistochemical analysis, and were sectioned at 70 nm and 100 nm thickness, respectively. D2 and D3 were tissue blocks from the hippocampus SR region without this phenotype, sectioned at 70 nm and 100 nm thickness, respectively.

\subsection{Manual annotation procedure}

vEM datasets were aligned using Fiji software TrakEM2\cite{ref56} plug-in. The ground\hy{}truth annotations were generated using Reconstruct or PyReconstruct\cite{ref57}. Astrocytes were identified based on characteristic ultrastructural features, including spongiform cytoplasm, the absence of synaptic inputs, and, when present, electron\hy{}dense cytoplasmic filaments. Annotators then traced membrane\hy{}enclosed cellular processes continuously across serial sections to reconstruct connected astrocytic processes throughout the volume. D1 and D2 were initially annotated by one annotator, and D3 and D4 by a second annotator. All four datasets were subsequently proofread by the same primary annotator, who was different from the initial annotator of each dataset and who checked the continuity of astrocytic processes across consecutive sections, excluded clearly non\hy{}astrocytic profiles, corrected disconnected regions, and ensured consistency with adjacent sections. Annotations judged ambiguous were additionally reviewed jointly by multiple annotators. The finalized annotations were subsequently exported as 2D TIFF masks using custom scripts for model training and evaluation.

\subsection{Overview of the vEM analysis and reconstruction pipeline}

Our lightweight and morphology\hy{}aware pipeline of astrocyte segmentation and reconstruction was designed as an astrocyte\hy{}centric framework to overcome the extreme spatial sparsity, complex branching, and the anisotropic resolution of vEM datasets. Instead of relying on computationally intensive 3D segmentation models, our framework leverages high\hy{}resolution 2D planar details and efficiently recovers 3D structures of astrocytes. The overall workflow consists of four integrated stages: (i) Foreground\hy{}aware patch sampling to address class imbalance (Fig.~\ref{fig:pipeline}A); (ii) Transformer\hy{}based 2D segmentation (Fig.~\ref{fig:pipeline}B) with sliding\hy{}window inference (Fig.~\ref{fig:pipeline}C) to extract astrocyte ultrastructures; (iii) Morphology\hy{}preserving 2D-to-3D hybrid post\hy{}processing (Fig.~\ref{fig:pipeline}D) to restore inter\hy{}section continuity and ensure volumetric coherence; (iv) Mesh reconstruction and Sholl analysis results (Fig.~\ref{fig:reconstruction}).

\subsection{Foreground\hy{}aware patch sampling from vEM data}

Foreground of astrocyte vEM data occupies only a small fraction of the tissue volume and appears as sparse, spatially fragmented features in 2D sections, resulting in severe class imbalance and background\hy{}dominated sampling. This challenge is further exacerbated by anisotropic voxel spacing. To address this, we employ a foreground\hy{}aware patch sampling strategy illustrated in Fig.~\ref{fig:pipeline}A. Fixed\hy{}size patches (224 $\times$ 224 pixels) are extracted from each 2D section, with patch centers sampled using a probabilistic mixture: 30\% from foreground pixels, and 70\% from the entire section. We also apply boundary\hy{}aware filtering to improve sample quality. Patches with more than 10\% low\hy{}intensity background are discarded, and patches consisting entirely of foreground or background are largely excluded, with 5\% retained to preserve extreme cases. This astrocyte\hy{}centric sampling strategy increases exposure to rare foreground and boundary regions where thin processes are most likely to be missed by random patch extraction.

\subsection{Transformer\hy{}based astrocyte segmentation}

Astrocyte segmentation is performed using a trans\-former\hy{}based U-Net architecture with a MiT-B2 (Mix Transformer) encoder (Fig.~\ref{fig:pipeline}B) originally introduced in SegFormer, which was initialized with ImageNet pretrained weights. The MiT encoder utilizes overlapping patch embeddings and a hierarchical multi\hy{}stage design, enabling progressive feature abstraction while maintaining spatial coherence. Its multi\hy{}scale representations capture both coarse somatic regions and fine distal astrocytic processes, facilitating segmentation of sparse and morphologically complex structures. Self\hy{}attention modules in transformer blocks model long\hy{}range spatial dependencies, integrating spatially distributed ultrastructure beyond the local convolutional receptive field. To improve computational efficiency on large vEM sections, attention is implemented with spatial reduction of key and value features in the first three stages. Subsequently, a U-Net--based decoder with skip connections and squeeze\hy{}and\hy{}excitation attention modules refines high\hy{}resolution features, improving boundary delineation of thin astrocytic processes.

Model training is performed on 224 $\times$ 224 patches extracted from individual vEM sections. Given the strong spatial continuity of astrocyte morphology along the z\hy{}axis, particular care is taken to avoid information leakage between training and validation data. Sections within each dataset are sequentially sorted and divided into 80\% training and 20\% validation subsets, ensuring that all patches from a given section were assigned exclusively to one split. This partition was applied only during training designated for each experiment (Table~\ref{tab:design}). Grayscale images are replicated across three channels to match the transformer encoder (Fig.~\ref{fig:pipeline}B), normalized using fixed mean and standard deviation values of {[}0.5, 0.5, 0.5{]}, and converted to PyTorch tensors. No additional random geometric or intensity data augmentation was applied during training.

For scalable inference on high\hy{}resolution vEM sections, we adopt an overlapping sliding\hy{}window strategy (Fig.~\ref{fig:pipeline}C). Each vEM section \(I_{z} \in \mathbb{R}^{X \times Y}\) is partitioned into 224 $\times$ 224 patches with 50\% overlap in both horizontal (\emph{x}) and vertical (\emph{y}) directions. As each pixel \((x,y)\) is covered by multiple patches, the aggregated probability map \(P_{z,c}(x,y)\) is computed as:

\begin{equation}
P_{z,c}(x,y) = \frac{1}{N(x,y)}\sum_{k=1}^{N(x,y)} p_{k,c}(x,y),
\label{eq:aggregate}
\end{equation}

where \(p_{k,c}(x,y)\) denotes the predicted probability for class \(c\) from patch \(k\) and \(N(x,y)\) is the number of overlapping patches covering pixel \((x,y)\). The final segmentation mask \(\hat{S}_{z} \in \{0,1\}^{X \times Y}\) is obtained by performing an \(argmax\) operation over the class dimension (i.e., selecting the index \(c\) that yields the highest aggregated probability)

\begin{equation}
\hat{S}_{z}(x,y) = \operatorname*{arg\,max}_{c} P_{z,c}(x,y).
\label{eq:argmax}
\end{equation}

This aggregation reduces boundary artifacts and preserves spatial continuity, while all predictions are performed at native resolution without rescaling. In the binary setting used in this paper, the maximum\hy{}probability assignment is equivalent to applying a 0.5 decision threshold to the foreground probability after softmax aggregation.

The training and inference are implemented in the PyTorch framework and executed on a single NVIDIA RTX 2060 GPU (12 GB VRAM). To accelerate the computational pipeline efficiently and optimize memory utilization, automatic mixed precision (AMP) is employed during training. Model parameters are optimized using the AdamW optimizer with an initial learning rate of \(6 \times 10^{- 5}\) and a weight decay of \(1 \times 10^{- 4}\). The network is trained for up to 50 epochs with a batch size of 64. To prevent overfitting, an early stopping mechanism with a patience of 5 epochs is applied based on the validation Dice score. To address class imbalance, training is supervised with a hybrid loss comprising equally weighted cross\hy{}entropy and Dice losses. All experiments were performed using a fixed random seed (2025) leveraging Python, NumPy, PyTorch, CUDA, and DataLoader workers.

The 2D segmentation experiments (Exp 1--3) used Python 3.11.14, PyTorch 2.5.1, torchvision 0.20.1, segmentation\hy{}models\hy{}pytorch 0.5.0, timm 1.0.24 and CUDA 12.1 on an NVIDIA RTX 2060 GPU with 12 GB of memory. The 2-fold and leave\hy{}one\hy{}volume\hy{}out models in Supplementary Table S1 used Python 3.11.5, PyTorch 2.13.0, torchvision 0.28.0, segmentation\hy{}models\hy{}pytorch 0.3.4, timm 0.9.7 and CUDA 13.0 on an NVIDIA A100-SXM4-40GB GPU of the Narval cluster of the Digital Research Alliance of Canada. The 3D nnU-Net comparison used nnunetv2 2.5.1 with Python 3.11.5, PyTorch 2.5.1, torchvision 0.20.1 and CUDA 13.0 on the same type of GPU. Manual annotation used Fiji (version 1.54t) with the TrakEM2 plugin, Reconstruct (version 1.1.0.0) and PyReconstruct (version 1.19.0), and Sholl analysis used the Fiji SNT plugin (version 5.0.2).

\subsection{Morphology\hy{}preserving post\hy{}processing}

Post\hy{}processing is a critical step for astrocyte segmentation due to the complexity of astrocyte morphological features and the reconstruction failure modes associated with thin, spatially fragmented processes. Although transformer\hy{}based 2D segmentation produces spatially coherent probabilistic maps, section-by\hy{}section predictions often inevitably contain fragmented components, spurious foreground regions, and discontinuities that hinder accurate 3D reconstruction. To enforce structural consistency across sections and preserve biologically meaningful astrocyte morphology, we implement an efficient, multi\hy{}stage, morphology\hy{}aware post\hy{}processing pipeline (Fig.~\ref{fig:pipeline}D) that operates collaboratively in both the 2D and 3D domains.

\emph{Soma\hy{}guided component refinement:} Section\hy{}wise segmentation lacks explicit 3D contextual information, leading to inconsistent false positives. For each section \(I_{z}\), we extract a set of foreground connected components \(\{C_{z,i}\}_{i = 1}^{N_{z}}\) from the predicted mask \(\hat{S}_{z}\). Each component \(C_{z,i}\) is characterized by its area \(A_{z,i}\) and centroid \(\mathbf{r}_{z,i} = \left( x_{z,i},\ y_{z,i} \right)\). To capture their cross\hy{}section distribution of these foreground components in physical space, centroids are embedded into 3D space as:

\begin{equation}
\mathbf{q}_{z,i} = \left( x_{z,i} \bullet s_{x},\ y_{z,i} \bullet s_{y},\ z \bullet s_{z} \right),
\label{eq:embed}
\end{equation}

where \(s_{x}\), \(s_{y}\) and \(s_{z}\) denote anisotropic voxel spacing along their respective axes. To account for size differences, each point is replicated proportionally to \(\sqrt{A_{z,i}}\). HDBSCAN\cite{ref58} clustering is then applied to identify spatially coherent components, and the cluster with the largest cumulative area is selected as the soma cluster. The actual soma\hy{}associated component in section \(I_{z}\) is identified as the foreground component situated closest to this reference coordinate. Let \(\mathbf{r}_{z}^{\mathrm{soma}}\) and \(A_{z}^{\mathrm{soma}}\) represent its centroid and area, respectively. The Euclidean distance to the estimated soma center is computed as

\begin{equation}
d_{z,i} = \left\| \mathbf{r}_{z,i} - \mathbf{r}_{z}^{\mathrm{soma}} \right\|_{2}.
\label{eq:dist}
\end{equation}

And components are removed only if they satisfy both size and spatial distance criteria:

\begin{equation}
A_{z,i} \geq \alpha A_{z}^{\mathrm{soma}}, \qquad d_{z,i} > \beta D_{z},
\label{eq:criteria}
\end{equation}

where \(D_{z}\) is the diagonal length of section \(I_{z}\), and \(\alpha\) and \(\beta\) are relative thresholds for area and distance, respectively. This soma\hy{}guided refinement suppresses large, spatially inconsistent connected components arising from the absence of volumetric context, while preserving soma\hy{}centered structures and their associated distal processes. By enforcing joint size and spatial distance constraints within a soma\hy{}centered reference frame, the procedure enhances volumetric coherence without introducing artificial morphological alterations.

\emph{Volumetric structural consolidation:} While soma\hy{}guided component refinement enforces cross\hy{}sectional consistency and removes prominent outliers, isolated false predictions may persist due to the inherently limited spatial context of section\hy{}wise segmentation. To further ensure global structural consistency, we perform volumetric connectivity analysis on the refined vEM stack. The segmentation is treated as a 3D binary volume, and unsupported single\hy{}section artifacts are first suppressed based on their lack of continuity across adjacent sections. Connected components are then identified in voxel space under 26-connectivity, and the k components with the largest volumetric support are retained as the principal astrocytic structures. This volumetric consolidation step removes residual disconnected fragments and segmentation artifacts that lack axial continuity and enforces global morphological coherence across the reconstructed astrocyte volume.

\emph{In\hy{}plane morphological correction:} The preceding soma\hy{}guided component refinement and volumetric consolidation effectively suppress spatially inconsistent false positives and noise, making the remaining cross\hy{}sectional masks highly confident and sufficiently clean. To further improve morphological plausibility, a minimal in\hy{}plane morphological operation is applied following volumetric consolidation. Specifically, a standard 2D binary hole\hy{}filling algorithm is independently employed on each section to eliminate small, artificial internal cavities caused by local segmentation inconsistencies. This final refinement ensures the solid topological structure of the reconstructed astrocyte without altering its external boundaries and distal processes.

\subsection{3D reconstruction and morphology analysis}

To facilitate high\hy{}fidelity visualization and qualitative structural assessment, the marching cubes algorithm\cite{ref59} implemented by mcubes (PyMCubes) library\cite{ref60} is applied to the stacked and post\hy{}processed 2D vEM sections, converting the resulting 3D volume into a continuous geometric surface model. This reconstruction process employs isosurface extraction to transform volumetric stacks of 2D binary masks into triangular mesh representations. With an intensity threshold of 0.5, a high\hy{}resolution geometric model was generated to capture the complex branching morphology of astrocytes. To control the computational complexity of grid generation for large\hy{}scale vEM datasets, we directly implement isosurface extraction based on the post\hy{}processed 2D stack (downsampled ten times on the xy\hy{}plane). The resulting surfaces are further processed and represented using PyVista\cite{ref61}. To preserve morphological fidelity, a stretch\hy{}and\hy{}scale transformation is applied to mesh vertices to compensate for anisotropic voxel resolution and any sampling factors introduced during segmentation.

\subsection{Evaluation metrics}

Dice, precision and recall are applied to quantitatively evaluate 2D segmentation performance. While these metrics provide a general evaluation of spatial overlap in plane, they exhibit two critical limitations in the context of astrocyte vEM segmentation. First, their volume\hy{}centric nature causes the large soma to dominate the score, potentially masking errors in thin astrocytic processes. Second, all false positives are penalized uniformly, despite distant artifacts being easily removed during post\hy{}processing.

To address this, we introduce a weighted Dice (wDice) as an astrocyte\hy{}centric complement to conventional Dice. Let \(G\) and \(P\) represent the binary ground\hy{}truth and predicted masks, respectively. We first compute the Euclidean distance transform map \(D(x,y)\), which measures the shortest distance from any spatial location \((x,y)\) to the nearest foreground pixel in \(G\). A spatial weight map \(W(x,y)\) is then defined using an exponential decay function:

\begin{equation}
W(x,y) = e^{-\gamma \cdot D(x,y)},
\label{eq:weight}
\end{equation}

where \(\gamma\) is the distance decay factor empirically set to 0.1. The wDice is subsequently defined as:

\begin{equation}
\resizebox{0.86\columnwidth}{!}{$\displaystyle
w\mathrm{Dice} = \frac{2\sum_{x,y} W(x,y)P(x,y)G(x,y)}{\sum_{x,y} W(x,y)P(x,y) + \sum_{x,y} W(x,y)G(x,y)},$}
\label{eq:wdice}
\end{equation}

which strategically ignores distant false positives while heavily penalizing false positives that distribute near the true astrocyte boundaries.

For the per\hy{}section error decomposition, we additionally report the false discovery rate (FDR), FDR = false positives / (false positives + true positives) = 1 $-$ precision, and the miss rate = false negatives / (false negatives + true positives) = 1 $-$ recall, with all counts taken as pixels within a section. In comparison with a 3D nnU-Net (Discussion), predictions and ground truth were evaluated on a common grid of 20 nm in\hy{}plane resolution. The native 5 nm ground\hy{}truth masks and final AstroSeg-vEM predictions from the complete pipeline were downsampled fourfold with a block-OR operation, whereby each 20 nm pixel was assigned as foreground if any of its 4 $\times$ 4 native pixels was foreground. This is the same operation applied to build the nnU-Net training data. The nnU-Net predictions are natively at 20 nm, and both models are scored against the same downsampled ground truth on all 83 D4 sections. Values on this grid are therefore not comparable with the native 5 nm values in Tables~\ref{tab:arch} and \ref{tab:ablation}.

\subsection{Statistics and reproducibility}

Unless otherwise noted, \emph{n} refers to the number of 2D sections evaluated within a single held\hy{}out volume per experimental condition, not independent volumes or animals. These sections are not independent biological replicates, as they derive from the same animal and tissue volume.

\section*{Data Availability}
The datasets generated and/or analyzed during the current study are not publicly available due to institutional policies but are available from the corresponding author on reasonable request.

\section*{Code Availability}
Source code is available under the GNU General Public License v3.0 (GPL-3.0) at \url{https://github.com/QuIN-Lab/AstroSeg-vEM}.

\section*{Acknowledgements}
B.D. and N.Y.K. acknowledge the support from Industry Canada and Natural Sciences and Engineering Research Council of Canada (Discovery Grant RGPIN-2025-04474). G.J.J., S.N., and J.Y.M. are supported by the KBRI basic research program through the Korea Brain Research Institute, funded by the Ministry of Science and ICT (26-BR-01-03, 26-BR-07-01) and the National Research Foundation of Korea (NRF, RS-2024-00419699).

\section*{Author Information}
\backhead{Contributions}
N.Y.K. and J.Y.M. conceived the idea, designed the research direction, and supervised the project. B.D. developed an automatic deep learning-based volume electron microscopy segmentation and reconstruction pipeline and performed related benchmark experiments. G.J.J. performed manual image analysis and S.N. acquired the electron microscopy images. B.D., J.Y.M., and N.Y.K. wrote the manuscript. All authors discussed the results and contributed to the paper.

\backhead{Corresponding author}
{\raggedright Correspondence and requests for materials should be addressed to: Na Young Kim (\href{mailto:nayoung.kim@uwaterloo.ca}{nayoung.kim@uwaterloo.ca}, 0000-0002-0290-7728), B. Deng (\href{mailto:b28deng@uwaterloo.ca}{b28deng@uwaterloo.ca}, 0000-0001-9625-829X), J.Y. Mun (\href{mailto:jymun@kbri.re.kr}{jymun@kbri.re.kr}, 0000-0003-0820-6233).\par}

\section*{Ethics declarations}
\backhead{Competing interests}
The authors declare no competing financial or non-financial interests.

\clearpage
\onecolumngrid
\setcounter{figure}{0}
\setcounter{table}{0}
\setcounter{equation}{0}
\renewcommand{\thefigure}{S\arabic{figure}}
\renewcommand{\thetable}{S\arabic{table}}
\renewcommand{\theHfigure}{S\arabic{figure}}
\renewcommand{\theHtable}{S\arabic{table}}
\makeatletter
\setlength{\@fptop}{0pt}\setlength{\@fpsep}{2.5em}\setlength{\@fpbot}{0pt plus 1fil}
\makeatother

\begin{center}
{\large Supplementary Material}\\[1.2em]
{\large\bfseries Lightweight Deep Learning Framework for Astrocyte-Centric 3D Reconstruction across Normal and Stress-associated Ultrastructural States}\\[1em]
Bowen Deng$^{1}$, Gwang-Jin Jeong$^{2,3}$, Seulgi Noh$^{2}$, Ji Young Mun$^{2}$ \& Na Young Kim$^{1,4}$
\end{center}
\vspace{1em}

\begin{figure}[htbp]
\centering
\includegraphics[width=0.95\textwidth]{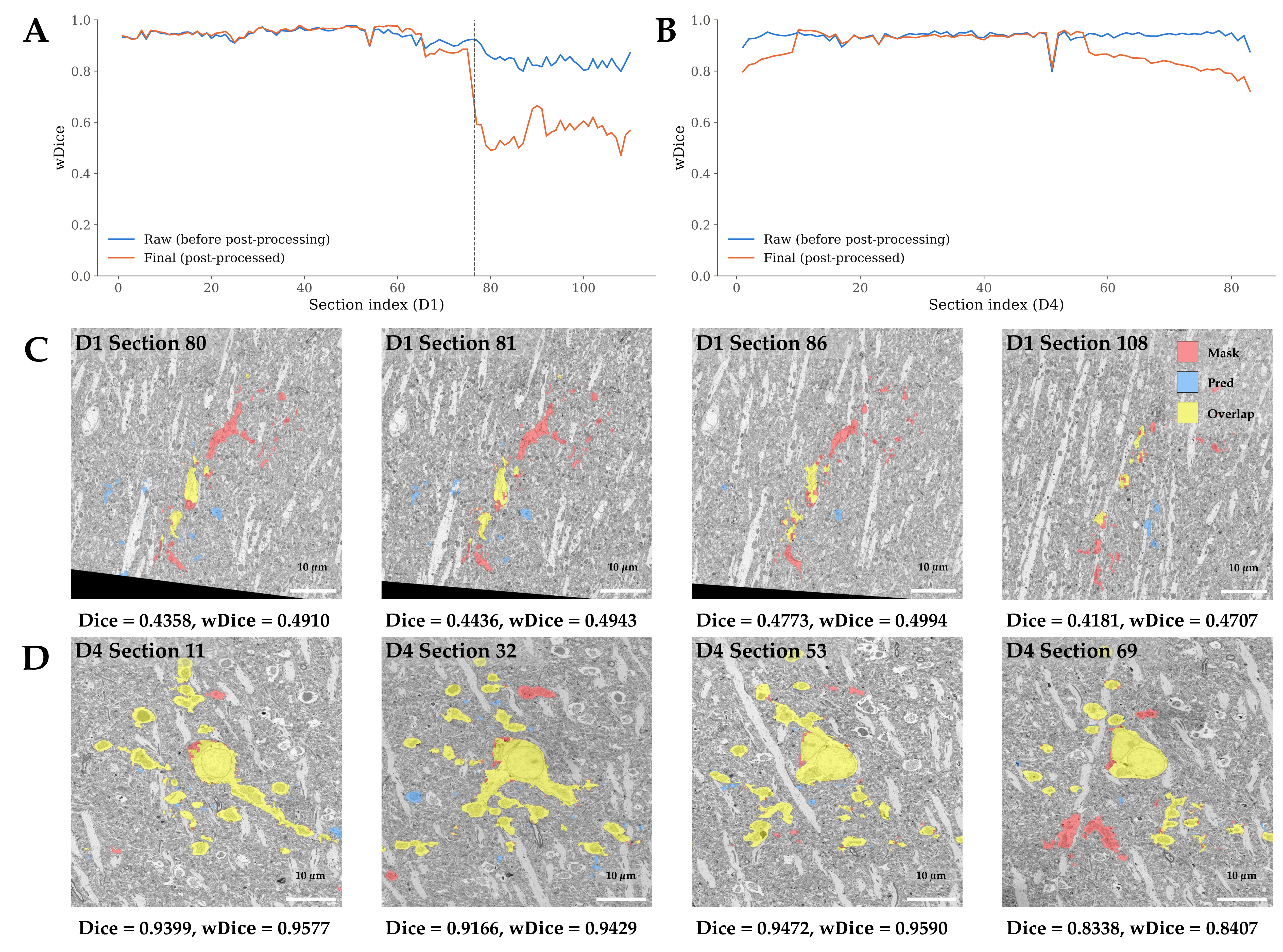}
\caption{\textbf{Comparison of per\hy{}section segmentation quality before and after post\hy{}processing across the D1 and D4 volumes.} \textbf{A} Per\hy{}section raw (before post\hy{}processing) and final (post\hy{}processed) wDice across the D1 volume. \textbf{B} Per\hy{}section raw and final wDice across the D4 volume, shown for comparison. \textbf{C} Overlays for the four post\hy{}processed outlier sections from D1. \textbf{D} Overlays for four representative sections from D4. Yellow = correctly predicted overlap, red = mask missed by the prediction, and blue = false positive prediction.}
\label{fig:S1}
\end{figure}

\begin{figure}[htbp]
\centering
\includegraphics[width=\textwidth]{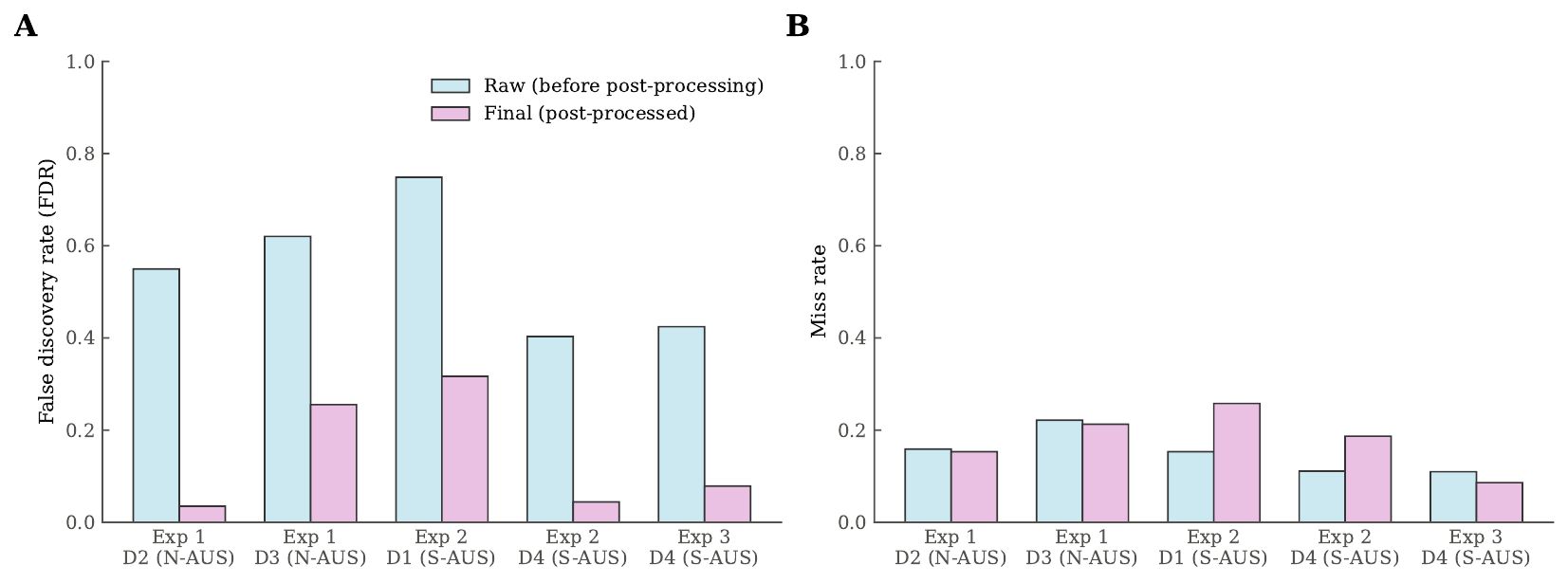}
\caption{\textbf{Post\hy{}processing error decomposition into false discovery rate and miss rate across experiments and datasets.} \textbf{A} False discovery rate (FDR = FP / (FP+TP) = 1 $-$ precision) before and after post\hy{}processing for each experiment and test dataset. \textbf{B} Miss rate (FN / (FN+TP) = 1 $-$ recall) before and after post\hy{}processing for each experiment and test dataset. The two rates have different denominators (predicted\hy{}positive and ground\hy{}truth\hy{}positive pixels) and are shown as complementary measures; bars show the mean across the sections of each held\hy{}out volume. TP, FP and FN are the numbers of true\hy{}positive, false\hy{}positive and false\hy{}negative pixels in a section.}
\label{fig:S2}
\end{figure}

\begin{figure}[htbp]
\centering
\includegraphics[width=0.95\textwidth]{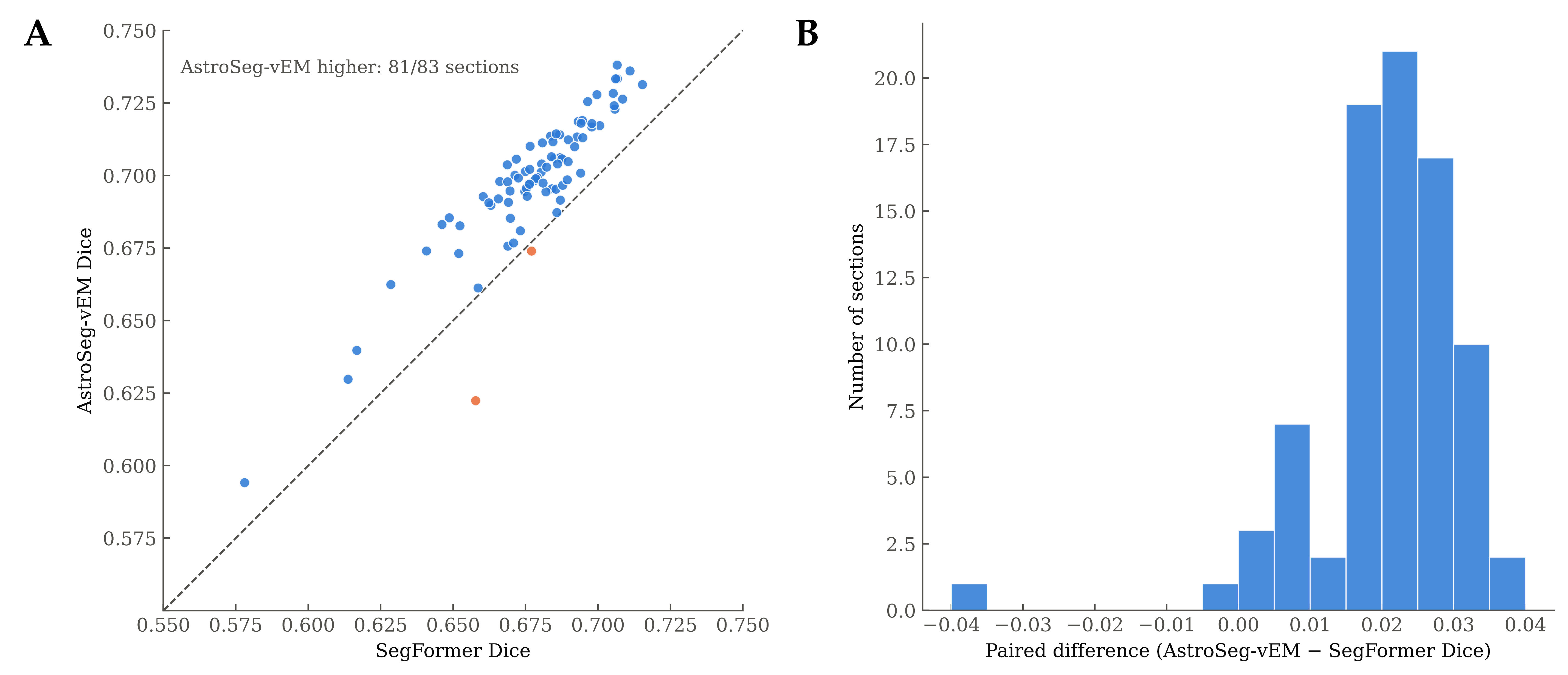}
\caption{\textbf{Paired, per\hy{}section comparison of AstroSeg-vEM and SegFormer Dice on Exp 3 D4.} \textbf{A} Per\hy{}section paired Dice (\emph{n} = 83 sections, same D4 volume evaluated by both models). Points above the diagonal are sections where AstroSeg-vEM scores higher. \textbf{B} Distribution of per\hy{}section paired differences (mean paired difference +0.021, moving\hy{}block bootstrap 95\% confidence interval 0.018--0.024 (block length 5--10 sections; 0.018--0.023 if serial dependence is ignored); positive in 81 of 83 sections; lag-1 autocorrelation of the paired differences 0.26, AR(1) effective sample size $\approx$ 49). This is a descriptive comparison within one volume; the sections are not independent replicates.}
\label{fig:S3}
\end{figure}

\begin{table}[htbp]
\caption{\textbf{Segmentation metrics before and after post\hy{}processing for all data\hy{}split protocols} Exp 1--3 are the main\hy{}text experiments, trained on the first 80\% and validated on the last 20\% of the sections of the listed volumes, whereas the other protocols use fewer training volumes. Values are mean $\pm$ standard deviation across the sections of the held\hy{}out test volume. Post\hy{}processed values for 2-fold-1 D2 vary strongly across sections.}
\label{tab:S1}
\begin{ruledtabular}
\renewcommand{\arraystretch}{1.25}
\begin{tabular}{lllcccccc}
& & & & & \multicolumn{2}{c}{\textbf{Raw}} & \multicolumn{2}{c}{\textbf{Post-processed}} \\
\cline{6-7}\cline{8-9}\noalign{\vskip 2pt}
\textbf{Protocol} & \textbf{Training} & \textbf{Validation} & \textbf{Test set} & \textbf{n} & \textbf{Dice} & \textbf{wDice} & \textbf{Dice} & \textbf{wDice} \\
\colrule\noalign{\vskip 3pt}
\multirow{2}{*}{Exp 1} & \multirow{2}{*}{D1, D4 (first 80\%)} & \multirow{2}{*}{D1, D4 (last 20\%)} & D2 (N-AUS) & 88 & 0.57 $\pm$ 0.19 & 0.91 $\pm$ 0.06 & 0.90 $\pm$ 0.06 & 0.91 $\pm$ 0.06 \\
 & & & D3 (N-AUS) & 53 & 0.50 $\pm$ 0.13 & 0.87 $\pm$ 0.05 & 0.76 $\pm$ 0.10 & 0.88 $\pm$ 0.05 \\
\noalign{\vskip 3pt}
\multirow{2}{*}{Exp 2} & \multirow{2}{*}{D2, D3 (first 80\%)} & \multirow{2}{*}{D2, D3 (last 20\%)} & D1 (S-AUS) & 110 & 0.38 $\pm$ 0.09 & 0.91 $\pm$ 0.05 & 0.68 $\pm$ 0.12 & 0.82 $\pm$ 0.18 \\
 & & & D4 (S-AUS) & 83 & 0.71 $\pm$ 0.02 & 0.94 $\pm$ 0.02 & 0.87 $\pm$ 0.05 & 0.89 $\pm$ 0.06 \\
\noalign{\vskip 3pt}
Exp 3 & D1--D3 (first 80\%) & D1--D3 (last 20\%) & D4 (S-AUS) & 83 & 0.70 $\pm$ 0.02 & 0.94 $\pm$ 0.02 & 0.92 $\pm$ 0.02 & 0.95 $\pm$ 0.02 \\
\noalign{\vskip 3pt}
\multirow{2}{*}{2-fold-1} & \multirow{2}{*}{D1} & \multirow{2}{*}{D4} & D2 (N-AUS) & 88 & 0.53 $\pm$ 0.18 & 0.94 $\pm$ 0.05 & 0.42 $\pm$ 0.43 & 0.47 $\pm$ 0.43 \\
 & & & D3 (N-AUS) & 53 & 0.20 $\pm$ 0.07 & 0.71 $\pm$ 0.10 & 0.46 $\pm$ 0.10 & 0.71 $\pm$ 0.12 \\
\noalign{\vskip 3pt}
\multirow{2}{*}{2-fold-2} & \multirow{2}{*}{D2} & \multirow{2}{*}{D3} & D1 (S-AUS) & 110 & 0.36 $\pm$ 0.10 & 0.92 $\pm$ 0.04 & 0.70 $\pm$ 0.06 & 0.89 $\pm$ 0.09 \\
 & & & D4 (S-AUS) & 83 & 0.64 $\pm$ 0.03 & 0.91 $\pm$ 0.02 & 0.89 $\pm$ 0.06 & 0.91 $\pm$ 0.06 \\
\noalign{\vskip 3pt}
LOVO fold 1 & D1, D2 & D3 & D4 (S-AUS) & 83 & 0.58 $\pm$ 0.04 & 0.95 $\pm$ 0.02 & 0.71 $\pm$ 0.04 & 0.92 $\pm$ 0.05 \\
\end{tabular}
\end{ruledtabular}
\end{table}

\begin{table}[htbp]
\caption{\textbf{Connected\hy{}component analysis of the Exp 3 D4 predictions before and after post\hy{}processing on the common 20 nm grid} Spurious components overlap no ground truth, missed components are ground\hy{}truth components without an overlapping prediction, and split and merged counts are ground\hy{}truth components covered by several predictions and predictions covering several ground\hy{}truth components. Values are mean $\pm$ standard deviation across the 83 D4 sections.}
\label{tab:S2}
\begin{ruledtabular}
\begin{tabular}{lccccc}
\textbf{Section set} & \textbf{\cstk{Connected components}{per section}} & \textbf{\cstk{Spurious}{components}} & \textbf{\cstk{Missed GT}{components}} & \textbf{\cstk{Split GT}{components}} & \textbf{\cstk{Merged predicted}{components}} \\
\colrule\noalign{\vskip 3pt}
Ground truth & 29.1 $\pm$ 2.7 & -- & -- & -- & -- \\
Raw prediction & 655.7 $\pm$ 39.1 & 599.9 $\pm$ 37.6 & 0.9 $\pm$ 1.0 & 10.0 $\pm$ 2.0 & 0.5 $\pm$ 0.7 \\
Post-processed prediction & 66.3 $\pm$ 17.7 & 30.3 $\pm$ 13.6 & 3.4 $\pm$ 2.7 & 5.0 $\pm$ 2.3 & 0.5 $\pm$ 0.7 \\
\end{tabular}
\end{ruledtabular}
\end{table}


\begin{thebibliography}{99}

\bibitem{ref1} Kol, A. et al. Astrocytes contribute to remote memory formation by modulating hippocampal--cortical communication during learning. Nat. Neurosci. 23, 1229--1239 (2020).

\bibitem{ref2} Zhou, Z. et al. Astrocytic cAMP modulates memory via synaptic plasticity. Proc. Natl. Acad. Sci. 118, e2016584118 (2021).

\bibitem{ref3} Wang, Y., Fu, A. K. Y. \& Ip, N. Y. Instructive roles of astrocytes in hippocampal synaptic plasticity: neuronal activity\hy{}dependent regulatory mechanisms. FEBS J. 289, 2202--2218 (2022).

\bibitem{ref4} Octeau, J. C. et al. An Optical Neuron-Astrocyte Proximity Assay at Synaptic Distance Scales. Neuron 98, 49--66.e9 (2018).

\bibitem{ref5} Ventura, R. \& Harris, K. M. Three\hy{}dimensional relationships between hippocampal synapses and astrocytes. J. Neurosci. 19, 6897--6906 (1999).

\bibitem{ref6} Witcher, M. R., Kirov, S. A. \& Harris, K. M. Plasticity of perisynaptic astroglia during synaptogenesis in the mature rat hippocampus. Glia 55, 13--23 (2007).

\bibitem{ref7} Kim, H. Y., Kim, S., Akaydin, A. N. et al. The rise of astrocytes: are they guardians or troublemakers of the brain disorder? Exp. Mol. Med. 58, 301--318 (2026).

\bibitem{ref8} Bushong, E. A., Martone, M. E., Jones, Y. Z. \& Ellisman, M. H. Protoplasmic Astrocytes in CA1 Stratum Radiatum Occupy Separate Anatomical Domains. J. Neurosci. 22, 183--192 (2002).

\bibitem{ref9} Verkhratsky, A. \& Nedergaard, M. Physiology of Astroglia. Physiol. Rev. 98, 239--389 (2018).

\bibitem{ref10} Verkhratsky, A., Butt, A., Li, B. et al. Astrocytes in human central nervous system diseases: a frontier for new therapies. Signal Transduct. Target. Ther. 8, 396 (2023).

\bibitem{ref11} Baldwin, K. T., Murai, K. K. \& Khakh, B. S. Astrocyte morphology. Trends Cell Biol. 34, 547--565 (2024).

\bibitem{ref12} Won, W., Bhalla, M., Lee, J.-H. \& Lee, C. J. Astrocytes as Key Regulators of Neural Signaling in Health and Disease. Annu. Rev. Neurosci. 48, 251--276 (2025).

\bibitem{ref13} Salmon, C. K. et al. Organizing principles of astrocytic nanoarchitecture in the mouse cerebral cortex. Curr. Biol. 33, 957--972.e5 (2023).

\bibitem{ref14} Schertel, A. et al. Cryo FIB-SEM: Volume imaging of cellular ultrastructure in native frozen specimens. J. Struct. Biol. 184, 355--360 (2013).

\bibitem{ref15} Špaček, J. Three\hy{}dimensional reconstructions of astroglia and oligodendroglia cells. Z. Zellforsch. 112, 430--442 (1970).

\bibitem{ref16} Witcher, M. R. et al. Three\hy{}dimensional relationships between perisynaptic astroglia and human hippocampal synapses. Glia 58, 572--587 (2010).

\bibitem{ref17} Aten, S. et al. Ultrastructural view of astrocyte arborization, astrocyte\hy{}astrocyte and astrocyte\hy{}synapse contacts, intracellular vesicle\hy{}like structures, and mitochondrial network. Prog. Neurobiol. 213, 102264 (2022).

\bibitem{ref18} Kikuchi, T. et al. Volume Electron Microscopy Study of the Relationship Between Synapses and Astrocytes in the Developing Rat Somatosensory Cortex. Cereb. Cortex 30, 3800--3819 (2020).

\bibitem{ref19} Chiappini, V. et al. Ultrastructure of astrocytes using volume electron microscopy: A scoping review. J. Physiol. 604, 1498--1517 (2026).

\bibitem{ref20} Shapson-Coe, A. et al. A petavoxel fragment of human cerebral cortex reconstructed at nanoscale resolution. Science 384, eadk4858 (2024).

\bibitem{ref21} Çiçek, Ö., Abdulkadir, A., Lienkamp, S. S., Brox, T. \&~Ronneberger, O. 3D U-Net: Learning Dense Volumetric Segmentation from Sparse Annotation. in Medical Image Computing and Computer-Assisted Intervention -- MICCAI 2016 (eds Ourselin, S., Joskowicz, L., Sabuncu, M. R., Unal, G. \& Wells, W.) vol. 9901 424--432 (Springer International Publishing, Cham, 2016).

\bibitem{ref22} Januszewski, M. et al. High\hy{}precision automated reconstruction of neurons with flood\hy{}filling networks. Nat. Methods 15, 605--610 (2018).

\bibitem{ref23} Lin, Z., Wei, D., Lichtman, J. \& Pfister, H. PyTorch Connectomics: A Scalable and Flexible Segmentation Framework for EM Connectomics. Preprint at \url{https://doi.org/10.48550/arXiv.2112.05754} (2021).

\bibitem{ref24} Isensee, F., Jaeger, P. F., Kohl, S. A., Petersen, J. \& Maier-Hein, K. H. nnU-Net: a self\hy{}configuring method for deep learning\hy{}based biomedical image segmentation. Nat. Methods 18, 203--211 (2021).

\bibitem{ref25} Yener, Y., Motta, A. \& Helmstaedter, M. Connectomic analysis of astrocyte\hy{}synapse interactions in the cerebral cortex. bioRxiv 2025.02.20.639274 (2025).

\bibitem{ref26} Dorkenwald, S., Li, P. H., Januszewski, M. et al. Multi\hy{}layered maps of neuropil with segmentation\hy{}guided contrastive learning. Nat. Methods 20, 2011--2020 (2023).

\bibitem{ref27} Sheridan, A. et al. Local shape descriptors for neuron segmentation. Nat. Methods 20, 295--303 (2023).

\bibitem{ref28} Ronneberger, O., Fischer, P. \& Brox, T. U-Net: Convolutional Networks for Biomedical Image Segmentation. in Medical Image Computing and Computer-Assisted Intervention -- MICCAI 2015 (eds Navab, N., Hornegger, J., Wells, W. M. \& Frangi, A. F.) vol. 9351 234--241 (Springer International Publishing, Cham, 2015).

\bibitem{ref29} Zhou, Z., Rahman Siddiquee, M. M., Tajbakhsh, N. \& Liang, J. UNet++: A Nested U-Net Architecture for Medical Image Segmentation. in Deep Learning in Medical Image Analysis and Multimodal Learning for Clinical Decision Support (eds Stoyanov, D. et al.) 3--11 (Springer International Publishing, Cham, 2018).

\bibitem{ref30} Oktay, O. et al. Attention U-Net: Learning Where to Look for the Pancreas. Preprint at \url{https://doi.org/10.48550/arXiv.1804.03999} (2018).

\bibitem{ref31} Alom, M. Z., Hasan, M., Yakopcic, C., Taha, T. M. \& Asari, V. K. Recurrent Residual Convolutional Neural Network based on U-Net (R2U-Net) for Medical Image Segmentation. Preprint at \url{https://doi.org/10.48550/arXiv.1802.06955} (2018).

\bibitem{ref32} Cao, Y., Liu, S., Peng, Y. \& Li, J. DenseUNet: densely connected UNet for electron microscopy image segmentation. IET Image Process. 14, 2682--2689 (2020).

\bibitem{ref33} Chen, L.-C., Papandreou, G., Kokkinos, I., Murphy, K. \& Yuille, A. L. Semantic Image Segmentation with Deep Convolutional Nets and Fully Connected CRFs. Preprint at \url{https://doi.org/10.48550/arXiv.1412.7062} (2016).

\bibitem{ref34} Chen, L.-C., Papandreou, G., Kokkinos, I., Murphy, K. \& Yuille, A. L. DeepLab: Semantic Image Segmentation with Deep Convolutional Nets, Atrous Convolution, and Fully Connected CRFs. IEEE Trans. Pattern Anal. Mach. Intell. 40, 834--848 (2018).

\bibitem{ref35} Chen, L.-C., Papandreou, G., Schroff, F. \& Adam, H. Rethinking Atrous Convolution for Semantic Image Segmentation. Preprint at \url{https://doi.org/10.48550/arXiv.1706.05587} (2017).

\bibitem{ref36} Chen, L.-C., Zhu, Y., Papandreou, G., Schroff, F. \& Adam, H. Encoder-Decoder with Atrous Separable Convolution for Semantic Image Segmentation. in Computer Vision -- ECCV 2018 (eds Ferrari, V., Hebert, M., Sminchisescu, C. \& Weiss, Y.) vol. 11211 833--851 (Springer International Publishing, Cham, 2018).

\bibitem{ref37} Cao, H. et al. Swin-Unet: Unet-Like Pure Transformer for Medical Image Segmentation. in Computer Vision -- ECCV 2022 Workshops (eds Karlinsky, L., Michaeli, T. \& Nishino, K.) vol. 13803 205--218 (Springer Nature Switzerland, Cham, 2023).

\bibitem{ref38} Zheng, S. et al. Rethinking semantic segmentation from a sequence-to\hy{}sequence perspective with transformers. in Proceedings of the IEEE/CVF conference on computer vision and pattern recognition 6877--6886 (2021).

\bibitem{ref39} Xie, E. et al. SegFormer: Simple and efficient design for semantic segmentation with transformers. Adv. Neural Inf. Process. Syst. 34, 12077--12090 (2021).

\bibitem{ref40} Chen, J. et al. TransUNet: Transformers Make Strong Encoders for Medical Image Segmentation. Preprint at \url{https://doi.org/10.48550/arXiv.2102.04306} (2021).

\bibitem{ref41} Wang, L. et al. UNetFormer: A UNet\hy{}like transformer for efficient semantic segmentation of remote sensing urban scene imagery. ISPRS J. Photogramm. Remote Sens. 190, 196--214 (2022).

\bibitem{ref42} Xing, C., Xie, R. \& Bader, G. D. RETINA: Reconstruction\hy{}based pre\hy{}trained enhanced TransUNet for electron microscopy segmentation on the CEM500K dataset. PLOS Comput. Biol. 21, e1013115 (2025).

\bibitem{ref43} Ciresan, D., Giusti, A., Gambardella, L. \& Schmidhuber, J. Deep neural networks segment neuronal membranes in electron microscopy images. Adv. Neural Inf. Process. Syst. 25 (2012).

\bibitem{ref44} Khadangi, A., Boudier, T. \& Rajagopal, V. EM\hy{}net: Deep learning for electron microscopy image segmentation. in 2020 25th international conference on pattern recognition (ICPR) 31--38 (IEEE, 2021).

\bibitem{ref45} Belevich, I. \&~Jokitalo, E. DeepMIB: User\hy{}friendly and open\hy{}source software for training of deep learning network for biological image segmentation. PLoS Comput. Biol. 17, e1008374 (2021).

\bibitem{ref46} Shah, A. et al. Automated image segmentation of scanning electron microscopy images of graphene using U-Net Neural Network. Mater. Today Commun. 35, 106127 (2023).

\bibitem{ref47} Nam, A. J. et al. Perisynaptic astroglial response to in vivo long\hy{}term potentiation and concurrent long\hy{}term depression in the hippocampal dentate gyrus. J. Neurosci. 45, e0943252025 (2025).

\bibitem{ref48} Li, H., Januszewski, M., Jain, V. \& Li, P. H. Neuronal Subcompartment Classification and Merge Error Correction. in Medical Image Computing and Computer Assisted Intervention -- MICCAI 2020 (eds Martel, A. L. et al.) 88--98 (Springer International Publishing, Cham, 2020).

\bibitem{ref49} Binley, K. E., Ng, W. S., Tribble, J. R., Song, B. \& Morgan, J. E. Sholl analysis: a quantitative comparison of semi\hy{}automated methods. J. Neurosci. Methods 225, 65--70 (2014).

\bibitem{ref50} Zhang, A. Q. et al. Acute systemic inflammation induces region\hy{}specific morphological remodeling of astrocytes and microglia concurrent with depression\hy{}like behavior. Histochem. Cell Biol. 163, 112 (2025).

\bibitem{ref51} Augustin, E. et al. Hippocampal Astrocyte Morphology Follows an Unexpected Trajectory with Age in a Transgenic Rodent Model of Tauopathy. Glia 73, 1502--1519 (2025).

\bibitem{ref52} Ortiz-Valladares, M. et al. Long\hy{}term intermittent fasting induces region- and sex\hy{}specific changes in astrocyte morphology and expression and anxiety\hy{}like behaviors in CD-1 mice. Behav. Brain Res. 494, 115745 (2025).

\bibitem{ref53} Viana, J. F. et al. Astrocyte structural heterogeneity in the mouse hippocampus. Glia 71, 1667--1682 (2023).

\bibitem{ref54} Longair, M. H., Baker, D. A. \& Armstrong, J. D. Simple Neurite Tracer: open source software for reconstruction, visualization and analysis of neuronal processes. Bioinformatics 27, 2453--2454 (2011).

\bibitem{ref55} Fok, A. H. K. et al. Deep learning segmentation with curvature consistency reveals astrocyte nanostructure across species. Cell Rep. 45, 117725 (2026).

\bibitem{ref56} Cardona, A., Saalfeld, S., Schindelin, J., Arganda-Carreras, I., Preibisch, S., Longair, M., Tomancak, P., Hartenstein, V. \& Douglas, R. J. TrakEM2 software for neural circuit reconstruction. PLoS One 7, e38011 (2012).

\bibitem{ref57} Chirillo, M. A., Falco, J. N., Musslewhite, M. D., Lindsey, L. F. \& Harris, K. M. PyReconstruct: A fully open\hy{}source, collaborative successor to Reconstruct. Proc. Natl. Acad. Sci. 122, e2505822122 (2025).

\bibitem{ref58} McInnes, L., Healy, J. \&~Astels, S. hdbscan: Hierarchical density based clustering. J. Open Source Softw. 2, 205 (2017).

\bibitem{ref59} Lorensen, W. E. \& Cline, H. E. Marching cubes: A high resolution 3D surface construction algorithm. ACM SIGGRAPH Comput. Graph. 21, 163--169 (1987).

\bibitem{ref60} Neila, P. M. PyMCubes: Marching cubes for Python. GitHub repository (2024).

\bibitem{ref61} Sullivan, B. \& Kaszynski, A. PyVista: 3D plotting and mesh analysis through a streamlined interface for the Visualization Toolkit (VTK). J. Open Source Softw. 4, 1450~(2019).

\end{thebibliography}
\end{document}